\documentclass[fleqn,10pt]{wlscirep}
\usepackage[utf8]{inputenc}
\usepackage[T1]{fontenc}

\newcommand{\nando}[2]{\textcolor{orange}{[NANDO: #2 ]}}

\definecolor{myblue}{RGB}{0, 128, 255}
\definecolor{mygreen}{rgb}{0.219, 0.462, 0.113}

\usepackage{amssymb}
\usepackage{amsmath,amsfonts,bm}

\def\eqref#1{equation~\ref{#1}}
\def\1{\bm{1}}

\DeclareMathAlphabet{\mathsfit}{\encodingdefault}{\sfdefault}{m}{sl}
\SetMathAlphabet{\mathsfit}{bold}{\encodingdefault}{\sfdefault}{bx}{n}

\DeclareMathOperator*{\argmin}{arg\,min}

\usepackage[normalem]{ulem}
\usepackage{float}
\usepackage{tabularx}
\usepackage{multirow}
\usepackage{colortbl} 
\usepackage{subcaption}
\usepackage{booktabs}
\usepackage{array}

\makeatletter
\g@addto@macro{\endtabular}{\rowfont{}}% Clear row font
\makeatother
\newcommand{\rowfonttype}{}% Current row font
\newcommand{\rowfont}[1]{% Set current row font
\gdef\rowfonttype{#1}#1\ignorespaces%
}
\makeatother
\usepackage{hyperref}

\title{Fine-grained Population Mapping from Coarse Census Counts and Open Geodata}

\author[1]{Nando Metzger*}
\author[2]{John E. Vargas-Mu\~{n}oz}
\author[1]{Rodrigo C. Daudt}
\author[3,2]{Benjamin Kellenberger}
\author[4]{Thao Ton-That Whelan}
\author[5]{Ferda Ofli}
\author[5]{Muhammad Imran}
\author[1]{Konrad Schindler}
\author[2]{Devis Tuia}

\affil[*]{Corresponding Author}
\affil[1]{ETH Zurich, Switzerland}
\affil[2]{\'{E}cole Polytechnique F\'{e}d\'{e}rale de Lausanne, Switzerland}
\affil[3]{Yale University, Untited States}
\affil[4]{International Committee of the Red Cross, Switzerland}
\affil[5]{Hamad Bin Khalifa University, Qatar}

\begin{abstract}
Fine-grained population maps are needed in several domains, like urban planning, environmental monitoring, public health, and humanitarian operations. Unfortunately, in many countries only aggregate census counts over large spatial units are collected, moreover, these are not always up-to-date. We present \textsc{Pomelo}, a deep learning model that employs coarse census counts and open geodata to estimate fine-grained population maps with 100$\,$m ground sampling distance. Moreover, the model can also estimate population numbers when no census counts at all are available, by generalizing across countries. In a series of experiments for several countries in sub-Saharan Africa, the maps produced with \textsc{Pomelo} are in good agreement with the most detailed available reference counts: disaggregation of coarse census counts reaches $R^2$ values of 85--89\%; unconstrained prediction in the absence of any counts reaches 48--69\%.
\end{abstract}

\begin{document}

\flushbottom
\maketitle

\thispagestyle{empty}

% Instructions from template:
% \noindent Please note: Abbreviations should be introduced at the first mention in the main text – no abbreviations lists. Suggested structure of main text (not enforced) is provided below.


\section*{Introduction}
\label{sec:introduction}

High-resolution population maps are crucial for many planning tasks, from urban planning~\cite{naess2001urban} to preparing humanitarian actions~\cite{lang2020earth} and effective disaster response~\cite{maas2019facebook}. Given the rapid population growth in many regions of the world~\cite{thomson2020gridded} and the increasing rate at which populations shift in response to environmental and social changes, it is important to maintain accurate, up-to-date maps. Unfortunately, census data are often only available at very coarse spatial resolution (e.g., one aggregate number for a district with hundreds or even thousands of km\textsuperscript{2}) and therefore not suitable as a basis for local planning: whether for sustainable land use and infrastructure management or for targeted disaster relief, planners need to know in more detail where the people are.
The problem is especially prevalent in developing countries in the global south, where humanitarian actions are more often needed yet census data availability and quality are limited.

Remote sensing products and other openly available geographical datasets like OpenStreetMap (OSM) can serve as auxiliary, high-resolution evidence to create fine-grained population density maps~\cite{vargas2020openstreetmap}. Yet, the design of effective population density models~\cite{stevens2015disaggregating} that combine such data sources with low-resolution census counts remains a challenge.
Generally speaking, two different approaches have been employed for population mapping~\cite{weber2018census}: bottom-up and top-down. 
Bottom-up methods~\cite{hillson2014methods, weber2018census} start from local surveys of population density, collected at a number of sample locations, and attempt to generalize from detailed but sparse samples to the unobserved regions to cover larger areas.
Researchers have proposed different ways to locally measure population density, such as counting the (average) number of people per rooftop area~\cite{hillson2014methods, weber2018census} or, if more resources are available for the local survey, specific average densities for different types of residential zones (urban-, rural-, and non-residential)~\cite{weber2018census, leasure2020national}.
A main drawback of bottom-up methods is that local surveys will necessarily remain extremely sparse and can hardly provide enough data points to scale population mapping up to the country level.
On the contrary, top-down approaches~\cite{stevens2015disaggregating,huang2021100} rely on census data, which ensures complete coverage at the expense of much lower spatial resolution, in some cases down to a single head count per large district. The task then becomes to disaggregate that data to a much finer resolution, often a regular grid. 
%-- often a regular grid -- with the help of dasymetric mapping methods.
%

Top-down approaches~\cite{stevens2015disaggregating, freire2016development} commonly use dasymetric disaggregation to redistribute the known, spatially coarse population counts for census areas on the order of many km\textsuperscript{2} across smaller spatial units~\cite{sapena2022empiric} -- for instance square blocks of size 100$\times$100 meters -- with the help of auxiliary variables that covary with population density. Examples of such covariates for population disaggregation are the presence of buildings~\cite{tiecke2017mapping}, building counts~\cite{huang2021100}, building volumes~\cite{ sridharan2013spatially}, and cumulative road lengths~\cite{reibel2005street}.
Dobson \textit{et al.}~\cite{dobson2000landscan} propose a weighted combination of several covariates, such as land cover and proximity to roads, to compute what portion of the population to assign to each target unit.
Stevens \textit{et al.}~\cite{stevens2015disaggregating} resort to machine learning and train a random forest model to predict population density from a set of covariates like building maps and night light images. 
%Note that for the top-down approach, only relative densities are needed, as the absolute scale is fixed by the aggregate number. 
%   
A limitation of the method proposed by Stevens \textit{et al.}~\cite{stevens2015disaggregating} is that administrative regions with known census counts are used as training units, which means that the input features (covariates) must be aggregated across all pixels in such a region; whereas at prediction (inference) time, density values are predicted for each individual pixel, respectively feature vector.
Consequently, the method's efficacy is reduced in countries with coarse census units, as aggregation over large spatial regions induces distribution shifts between the training and test data.

Recent work has employed deep learning, in particular convolutional neural networks (CNNs~\cite{lecun2015deep}) for population estimation, in an attempt to better account for spatial context. Gervasoni \textit{et al.}~\cite{gervasoni2018convolutional} used a CNN to map data extracted from OSM to disaggregation weights per pixel with 200$\times$200$\,$m ground sampling distance (GSD). This allows one to super-resolve census counts to a finer granularity; however, since the predicted weights are relative, the method cannot be used to predict population numbers for regions without census data. Jacobs \textit{et al.}~\cite{jacobs2018weakly} trained a CNN to predict fine-grained population maps from very high resolution (VHR) optical satellite imagery and census data. This is viable for individual cities, but difficult to scale up to larger geographic contexts, due to the limited availability and high cost of VHR images.
Both CNN-based methods~\cite{gervasoni2018convolutional,jacobs2018weakly} depend on fairly fine-grained initial counts with census units of few km\textsuperscript{2}, e.g., urban census blocks in France, respectively the USA. For comparison, the administrative units for which census data are available in much of Africa have sizes on the order of several hundred to several thousand km\textsuperscript{2}, which precludes the use of such methods.

Here, we propose a methodology to estimate fine-grained population maps from such very coarse census data by fusing them with covariate maps of higher resolution.
As inputs, we use publicly available products derived from remote sensing images and other open data sources, e.g., building footprint maps, night light images, and OSM road layers. They are used to train a model that predicts population on a spatial grid with 100$\times$100$\,$m GSD.
Our method, which we call \textsc{Pomelo} (short for ``\underline{po}pulation \underline{m}apping by \underline{e}stimation of \underline{l}ocal \underline{o}ccupancy rates''), is inspired by work on \emph{guided super-resolution} of low-resolution images~\cite{diebel2006nips,hui2016depth,Lutio_2019_ICCV}. % \cite{ham2017robust}
%In that work, the authors fit a neural network using an aggregating loss function to increase the resolution of depth maps by embedding information of high-resolution RGB images.
%
In an experimental evaluation with data from three different countries in Sub-Saharan Africa (Tanzania, Zambia, Mozambique) \textsc{Pomelo} delivers significantly more accurate population maps compared to several baseline methods, including the pioneering work of Stevens \textit{et al.}~\cite{stevens2015disaggregating} with which the widely used WorldPop maps~\cite{worlpop2022} are created.
Moreover, \textsc{Pomelo} can not only disaggregate existing census data, but can also predict population maps in the absence of census counts. Consequently, it can be deployed to regions or countries where no suitable census information is available -- of course these estimates, inevitably, have higher uncertainties, as the total population is no longer constrained by a known aggregate number. In summary, \textsc{Pomelo} provides knowledge about where people are at the hectometer scale, even if census data are not available.

%\textcolor{orange}{KS: do we need the following paragraph? I always try to avoid it, the information content of such a list is $\approx$0 bit...}

%The paper is organized as follows: In Section~\ref{data} we describe the data used as input of our proposed model. Section~\ref{Method} presents our proposed method, \textsc{Pomelo}. Section~\ref{Experiments} shows the experimental setup, and results, using three different evaluation strategies, while the results are discussed in Section~\ref{sec:discusssion}. Finally, Section~\ref{Conclusion} presents our conclusions.

\section*{Results}
\label{sec:results}
%\subsection*{Evaluation strategies}
%\label{subsec:eval_strategy}

%We evaluate the performance of \textsc{Pomelo} and other state-of-the-art baseline models using three different evaluation strategies. In all three strategies, the estimated maps are evaluated by aggregating the per-pixel counts back to a list of $n_c$ population numbers $\mathbf{\hat{c}}$ at finest available census level and comparing them to the actual census counts $\mathbf{c}$ in terms of $R^2$, mean absolute error (MAE), and mean absolute percentage error (MAPE):
We compare \textsc{Pomelo} with other state-of-the-art baseline models using three different evaluation strategies, and three performance metrics (see Methods section), namely $R^2$, mean absolute error (MAE), and mean absolute percentage error (MAPE):

\vspace{1em}
\noindent
\underline{\itshape Coarse supervision:} In this strategy, used by several works in the literature~\cite{stevens2015disaggregating,mennis2003generating,grippa2019improving}, the census data at the finest available level are reserved exclusively for performance evaluation. First, the available census data are artificially coarsened by aggregating them to the next-higher (``coarse'') level of administrative regions (e.g., counts at district level are aggregated into a single count per province). Then, the coarse level regions are divided into five random folds, of which three serve as training set, one as validation set for hyper-parameter tuning and checkpointing the model on the best MAPE, and one as test set to measure the model's predictive skill. Finally, the model is trained to predict gridded population numbers at 100~m GSD, such that they add up to the coarse level counts. To measure performance, the gridded population estimates are aggregated to obtain population numbers for each \emph{fine} level administrative region in the test set and are then compared to the corresponding census data.

This scheme corresponds to an application scenario where fine-level census data are not available: all model fitting and mapping is only based on the coarse level. The fine level counts are employed exclusively to evaluate prediction performance.

\begin{figure*}[tb]
\centering
\begin{subfigure}[h]{0.29\linewidth}
\includegraphics[height=6cm]{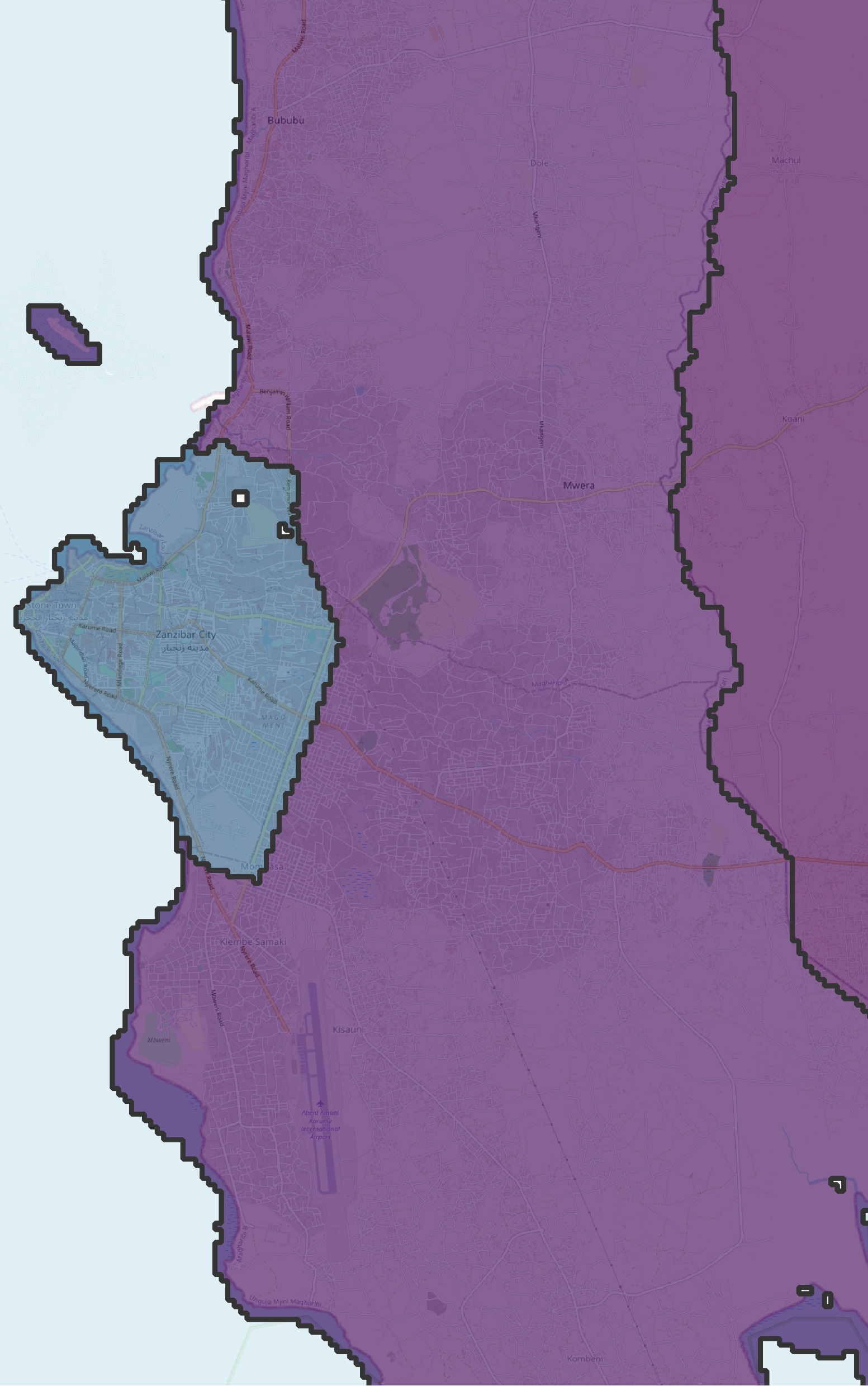}
\caption{Coarse (district level)\\\,}
\end{subfigure} 
\begin{subfigure}[h]{0.29\linewidth}
\includegraphics[height=6cm]{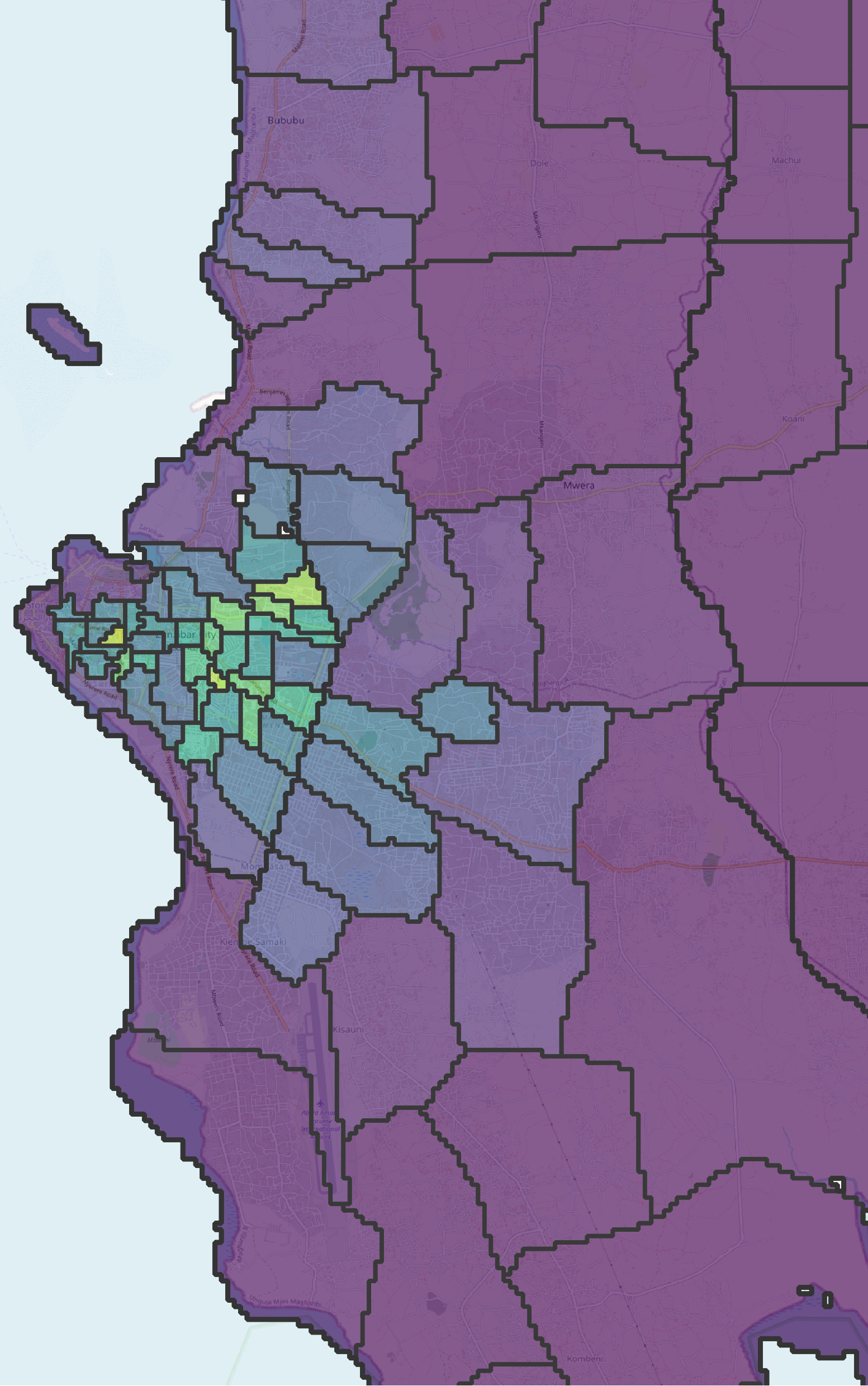}
\caption{Fine (ward level)\\\,}
\label{fig:census_data_overview_b}
\end{subfigure}% 
\begin{subfigure}[h]{0.29\linewidth}
\includegraphics[height=6cm]{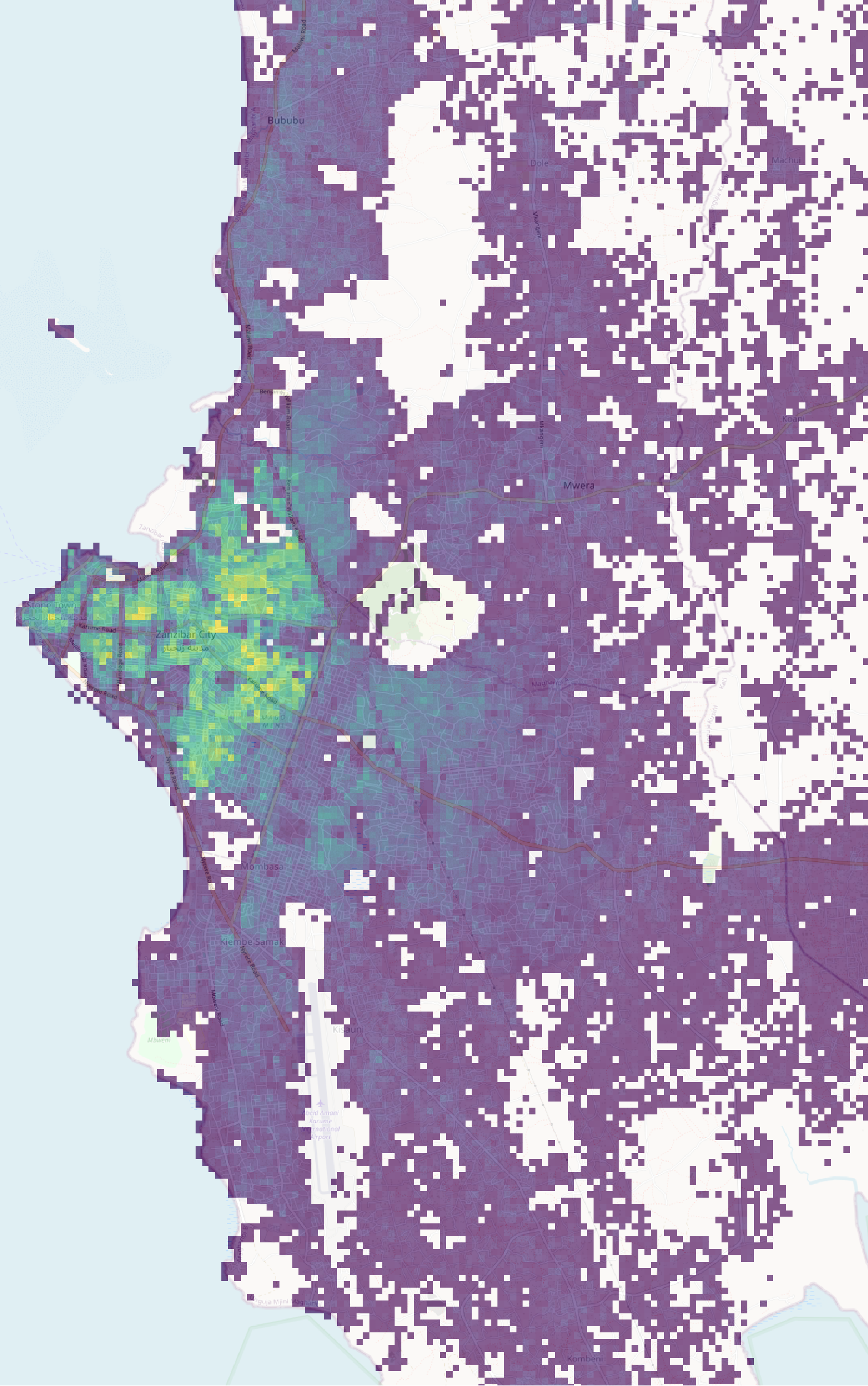}
\caption{Pixel (prediction level);\\ 100~m resolution}
\end{subfigure}%
\begin{subfigure}[h]{0.12\linewidth}
\includegraphics[height=4cm]{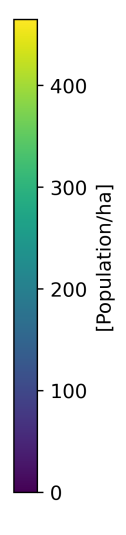}
%\caption{cbar}
\end{subfigure}%
\caption{Overview of the different level of population data. Shown location: Zanzibar City, Tanzania. The map visualizations were created in QGIS 3.14 \cite{QGIS_software}.}
\label{fig:census_data_overview}
\end{figure*}

The entire training and evaluation procedure is run as a five-fold cross-validation, rotating the validation and test folds.
Figure~\ref{fig:census_data_overview} depicts the coarse census data used for training, the fine level census data used for evaluation, and the fine-grained population maps obtained by our model, for the surroundings of Zanzibar, Tanzania.

\vspace{1em}
\noindent
\underline{\itshape Fine supervision:} In the second evaluation scheme, census data at the finest available level are used for supervision. The data are \emph{not} coarsened, rather the original fine-level data are divided into five random folds. These folds are split along the same coarse-level administrative boundaries as above, such that the training, validation and test portions during cross-validation are identical to the coarse supervision scenario. The model is trained to predict gridded population numbers at 100~m GSD, such that they add up to the fine-level counts. Just like above, the gridded population estimates are aggregated to obtain population numbers for each fine-level administrative region in the held-out test fold (e.g., Figure~\ref{fig:census_data_overview_b}), which are then compared to the corresponding census data to compute performance metrics.

This scheme corresponds to the actual, realistic population mapping task with the available census granularity. The finest available resolution is used to learn the best possible model, geographically separating training and testing regions to ensure an unbiased evaluation.
Note that the combined test set, after holding out each fold once for testing, is the same for \textit{coarse supervision} and \textit{fine supervision}, such that the error metrics are directly comparable.

\vspace{1em}
\noindent
\underline{\itshape Transfer task:} Besides census disaggregation, \textsc{Pomelo} can also be used to estimate population density for a given country in the absence of census numbers. To evaluate the performance on such more challenging task, we use data from seven different countries. One of the countries is held out as test set, and the data from the remaining six countries are randomly split into 80\% for training and 20\% for validation (i.e., hyper-parameter tuning). Then, the model is trained to predict gridded population numbers at 100~m GSD, using the \emph{fine-level} strategy.
Finally, the model is deployed for the held-out country, and evaluated in the same way as above.

For each test country we run five models with different random parameter initialization and train/validation splits and report the average and standard deviation of each performance metric. 
We compare our proposed method with other four baseline methods, described in the Methods section: Building count disaggregation, the random forest model used by WorldPop \cite{stevens2015disaggregating}, a Markov Random Field based method for population disaggregation, and a Convolutional Neural Network model. 

\vspace{1em}
For evaluation, we use data from Tanzania and Zambia with the three aforementioned evaluation strategies.
Mozambique is used only in the \textit{transfer task} evaluation.
The data available for that country is not suitable for the other two evaluation schemes: in Mozambique there are only 413 fine-level regions grouped into 156 coarse-level regions. The low aggregation factor of only $2.6\times$ causes the performance metrics to saturate at a similar value of 83\% $R^2$ for all methods, while giving little indication about the correctness of the pixel-level maps. 
% \john{}{TODO: Mention briefly the other baselines}

\subsection*{Results -- coarse supervision} \label{sec:c_train}
Table~\ref{tab:coarse_task_all} presents the performance metrics for the coarse supervision strategy in Tanzania and Zambia.
%The small scaling factor of Mozambique of only 2.6x leads to very saturated scores for all methods, while also diminishing the expressiveness about the validity of the pixel-level maps. We found the dataset of Mozambique to be unsuitable the performance evaluation of tasks where the predictions are constrained by the coarse level census.
%($\approx$100 coarse administrative districts with census counts).
The first two rows show learning-free methods that are only capable of pure disaggregation, i.e., they cannot estimate population numbers for regions without known (aggregate) census counts. The MRF formulation brings a marked improvement compared to the simple disaggregation scheme based only on building counts.
For the three learning-based methods, we first predict population numbers without using the target regions' census counts, then normalize those counts to relative fractions and use those as weights for dasymetric disaggregation.
\textsc{Pomelo} achieves the best performance in all three metrics, with the closest competitor being the MRF. An interesting observation is that the region-based RF method~\cite{stevens2015disaggregating}, when trained with such coarse census data, does not even reach the performance of simple building count disaggregation. The evaluation results for Zambia largely mirror those for Tanzania, where the MRF is the closest to \textsc{Pomelo} in terms of $R^2$ and MAE. Next follows the CNN, which handles Zambia much better than Tanzania. %, \nando{}{presumably due to the much lower scale difference: in the former a coarse region is, on average, 9.5× larger than a fine one, in the latter the factor is 21.5×.}

\begin{table}[t]
    \caption{Performance with coarse supervision for Tanzania and Zambia.}
    \label{tab:coarse_task_all}
    \centering
    %\scriptsize
    \begin{tabular}{@{}lllll@{}}
        \toprule
        Evaluation Set & Method   & \multicolumn{1}{c}{$R^2 \uparrow$ [\%]}  & \multicolumn{1}{c}{MAE $\downarrow$} & \multicolumn{1}{c}{$\!\!$MAPE $\downarrow$ [\%]}
        \\ \midrule 
        \multirow{5}{*}{Tanzania}
        & Building disaggregation                     & 65.2 & 3'700 & 23.1  \\ 
            & MRF                   & 78.6 &  3'300  & 21.7 \\ 
            \cline{2-5}
            & RF per region \cite{stevens2015disaggregating}      & 49.0 $\pm$ 3.0 & 4'380 $\pm$  \textcolor{white}{0}50 & 26.9 $\pm$ 0.2 \\
            & CNN                     & 69.5 $\pm$ 2.0 & 4'200 $\pm$ 100 & 28.8 $\pm$ 0.5  \\
            & \textsc{Pomelo}                                           & 85.7 $\pm$ 0.9  & 3'100 $\pm$ \textcolor{white}{0}40 & 21.6 $\pm$ 0.2  \\
        \midrule
        \multirow{5}{*}{Zambia}
            & Building disaggregation          & 82.1  & 4'300 & 53.9  \\ 
            & MRF                              & 87.0  &  3'900 & 51.3 \\ 
            \cline{2-5} 
            & RF per region \cite{stevens2015disaggregating}                 & 76.4 $\pm$ 0.0 & 4'750 $\pm$ \textcolor{white}{0}90 & 60.9 $\pm$ 1.2\\ 
            & CNN                     & 83.9 $\pm$ 0.7 & 4'250 $\pm$ \textcolor{white}{0}40 &  51.4 $\pm$ 1.7 \\   
            & \textsc{Pomelo}         & 87.9 $\pm$ 0.1 & 3'730 $\pm$ \textcolor{white}{0}30 &  50.4 $\pm$ 0.4 \\
        \bottomrule
    \end{tabular}
\end{table}

\subsection*{Results -- fine supervision}

Table~\ref{tab:fine_task_all} shows quantitative performance when the models are trained and evaluated on the finer census level.
Note that we cannot use the learning-free building disaggregation and MRF baselines with this evaluation strategy, since they are only designed for disaggregation of regions with known overall population numbers (which makes no sense in the fine supervision setting, as the target quantity would have to be known in advance).
For Tanzania, we again observe the best performance with \textsc{Pomelo}, although the advantage is smaller than in the case of coarse supervision (Table~\ref{tab:coarse_task_all}).
%(Table~\ref{tab:coarse_res_tza}).  
The performance of the region-based RF improves a lot when trained with fine supervision, showing that the coarse supervision does not offer enough supervision signal for training at the region level.
The second section of Table~\ref{tab:fine_task_all} shows results for Zambia. \textsc{Pomelo} maintains a slight edge in terms of $R^2$ and absolute error. It does not fare as well in terms of MAPE, which is due to large relative errors in few, sparsely populated regions (that nonetheless translate to small errors of the absolute population count): just by removing Zambia's smallest census region with only 28 inhabitants from the evaluation, the MAPE of \textsc{Pomelo} drops to 45\%.
%\nando{}{As such, we observe that the MAPE score is not a robust when applied to. When simply removing the Zambia's smallest census region (census count: 28, \textsc{POMELO}'s prediction: 1243) from the evaluation set, the MAPE score improves by 3 percentage points in favor of \textsc{POMELO}.}
%We observe that by removing the Zambia's smallest census region from the evaluation set, the error measured by MAPE is reduced by 3 percentage points in favor of \textsc{Pomelo}.

\begin{figure*}[t]
    \begin{subfigure}[h]{0.48\linewidth}
        \includegraphics[width=\linewidth]{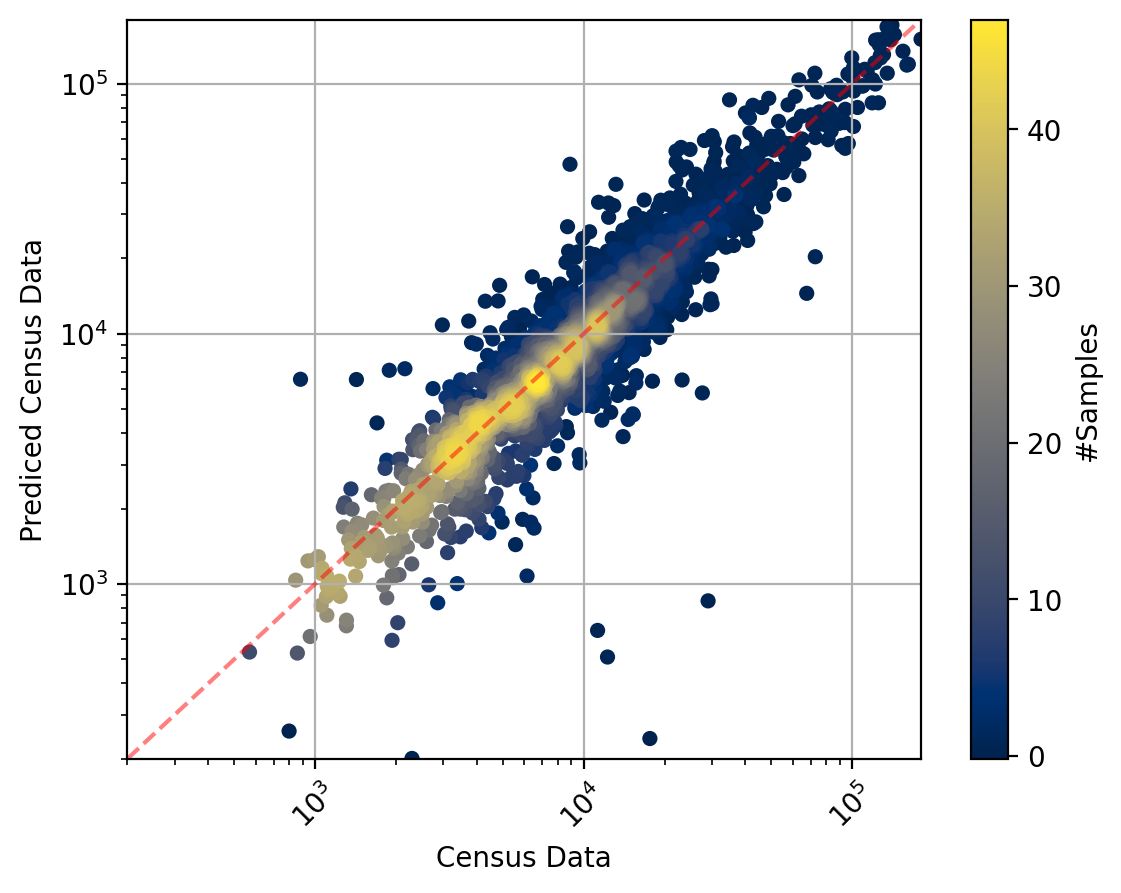}
        \caption{Tanzania}
    \end{subfigure}
    \hfill
    \begin{subfigure}[h]{0.48\linewidth}
        \includegraphics[width=\linewidth]{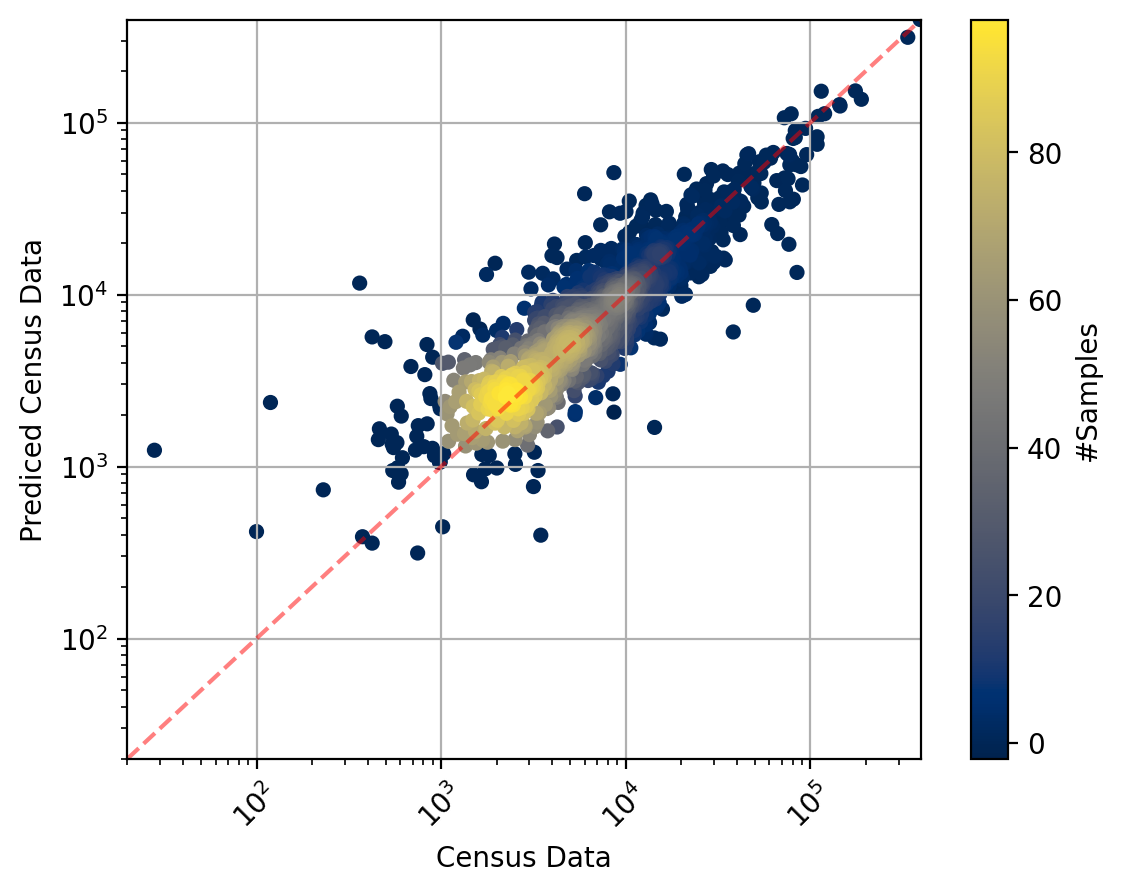}
        \caption{Zambia}
    \end{subfigure}%
    \caption{Comparison between the estimates predicted by \textsc{Pomelo} and the respective ground truth census in the fine level training task.} 
    %\john{}{Take a look why there is some point with almost zero census population. Should we redo the scatterplot?} \nando{}{I don't think we should remove it. it is what it is: the census is 0 and out model made the mistake to predict population there. If a reviewer complains we can always explain that there where buildings detected in the region, they are only for agriculture.}}
    \label{tab:scatter_plot_fine_level}
\end{figure*}

Figure~\ref{tab:scatter_plot_fine_level} shows scatter plots (census counts versus our predicted counts) for the fine-level administrative regions of both Tanzania and Zambia. In both cases, the data are, with few exceptions, close to the (red, dashed) diagonal that corresponds to the ideal result, where predictions and ground truth coincide.

%
%The Random Forest, on the contrary, performs better for Zambia, as the smaller scaling factor (9.5$\times$, compared to 21.5$\times$ for Tanzania) reduces the domain shift between the training and testing units. \nando{}{We need to rewrite this last part, the domain shift is between fine-pixel, not fine-coarse!!}

\begin{table}[t]
    \caption{Performance with fine supervision for Tanzania and Zambia.}
    \label{tab:fine_task_all}
    \centering
    %\scriptsize
    \begin{tabular}{@{}lllll@{}}
        \toprule
        Evaluation Set & Method   & \multicolumn{1}{c}{$R^2 \uparrow$ [\%]}  & \multicolumn{1}{c}{MAE $\downarrow$} & \multicolumn{1}{c}{$\!\!$MAPE $\downarrow$ [\%]}
        \\ \midrule 
        \multirow{3}{*}{Tanzania} 
            & RF per region \cite{stevens2015disaggregating}         & 79.1 $\pm$ 2.6  & 3320 $\pm$ 70 & 22.1 $\pm$ 0.3 \\
            & CNN                      & 80.8 $\pm$ 2.1  & 3290 $\pm$ 80 & 21.7 $\pm$ 0.4 \\  
            & \textsc{Pomelo}                                               & 87.6 $\pm$ 0.2 & 2890 $\pm$ 20 & 20.4 $\pm$ 0.3 \\ 
        \midrule
        \multirow{3}{*}{Zambia} & RF per region \cite{stevens2015disaggregating}        & 87.6 $\pm$ 0.0 & 3740 $\pm$ 40 & 46.5 $\pm$ 0.4 \\ 
            & CNN                      & 88.2 $\pm$ 0.4 & 3680 $\pm$ 50 & 44.9 $\pm$ 0.4 \\ 
            & \textsc{Pomelo}           & 88.7 $\pm$ 0.3 & 3650 $\pm$ 20 & 48.0 $\pm$ 0.6  \\  
        \bottomrule
    \end{tabular}
\end{table}

\subsection*{Transfer Task}

We also evaluate \textsc{Pomelo} in a scenario where no census data are available for the target country. Here, the models must instead be trained on other countries and should be able to generalize to a potentially different unknown geographical context. From the seven countries considered (Tanzania, Zambia, Mozambique, Rwanda, Uganda, Democratic Republic of Congo, and Nigeria), we select six for training and validation and report the evaluation metrics on the respective held out country (Tanzania, Zambia, and Mozambique) in Table~\ref{tab:transfer_task_all}.

% For technical reasons we could not train the CNN baseline for this experiment, because the large training regions -- especially in the Democratic Republic of Congo -- would have required 40 Gigabytes of GPU memory, substantially more than we had available.
%
The first rows of each section in the table (``average occupancy rate'') show the results of a na\"\i ve baseline that computes the average occupancy rate in the six training countries and multiplies it with the building count map of the target country to obtain the population map. One can see that this simple approach does not perform well in any of the countries, barely reaching a positive $R^2$ score only in Zambia.
The region-based RF model does not generalize well either, to the point that the $R^2$ metric becomes negative and the MAPE exceeds 100\% for Zambia and Mozambique.
The performance of the CNN is still rather good across all evaluation settings, despite the more challenging scenario. \textsc{Pomelo} exhibits even better overall performance, with the exception of a slightly higher MAE in Mozambique. 

As a proxy for a ``general'' model that is valid for an entire geographic area, we also create a training set with data from all seven countries (including the respective target country), and apply the resulting model to the (unseen) test set of the target country, using five-fold cross-validation to cover the whole country.
For comparison, we also run models trained specifically for each country, but without dasymetric rescaling, to keep the comparison fair. That scenario would normally not be relevant in practice, since the training implies that aggregate counts are available except in the special situation that no current census counts are available and a previously trained model is reused. %
However, by removing the influence of the postprocessing it shows the ability of the country-specific model to predict unconstrained, absolute numbers.
As expected, the country-specific models perform better than the one learned only with data from four other countries, but not as well as a model trained on all seven countries, presumably due to the larger amount and variability of training samples.

\begin{table}[t]
    \caption{Performance on the transfer task for all countries.}
    \label{tab:transfer_task_all}
    \centering
    %\scriptsize
    \begin{tabular}{@{}lllll@{}}
        \toprule
        Evaluation Set & Method   & \multicolumn{1}{c}{$R^2 \uparrow$ [\%]}  & \multicolumn{1}{c}{MAE $\downarrow$} & \multicolumn{1}{c}{$\!\!$MAPE $\downarrow$ [\%]}
        \\ \midrule
        \multirow{5}{*}{Tanzania} & Average occupancy rate      &     \textcolor{white}{0}-52.9 \textcolor{white}{$\pm$ 00.0}  &     13'200 \textcolor{white}{$\pm$ 0'000} & \textcolor{white}{0}98.0 \textcolor{white}{$\pm$ 00.0}\\
         & RF per region \cite{stevens2015disaggregating}        & \textcolor{white}{-0}22.1 $\pm$ \textcolor{white}{0}10.1  & \textcolor{white}{0}8'000 $\pm$ \textcolor{white}{0'}700  & \textcolor{white}{0}66.5 $\pm$ \textcolor{white}{00}6.2\\
         & CNN                                                  & \textcolor{white}{-0}34.6 $\pm$ \textcolor{white}{03}6.2  & \textcolor{white}{0}6'800 $\pm$ \textcolor{white}{0'}300 & \textcolor{white}{0}49.6 $\pm$ \textcolor{white}{00}3.0\\ 
         & \textsc{Pomelo}                                      & \textcolor{white}{-0}68.1 $\pm$ \textcolor{white}{03}0.9  & \textcolor{white}{0}5'200 $\pm$ \textcolor{white}{0'}200 & \textcolor{white}{0}35.6 $\pm$ \textcolor{white}{00}1.0\\ 
        \cline{2-5}
         & \textsc{Pomelo} -- 7 countries                      & \textcolor{white}{-0}81.7 $\pm$ \textcolor{white}{03}0.6  & \textcolor{white}{0}3'170 $\pm$ \textcolor{white}{0'0}20 & \textcolor{white}{0}21.7 $\pm$ \textcolor{white}{00}0.6\\ 
         & \textsc{Pomelo} -- Tanzania only                    & \textcolor{white}{-0}82.0 $\pm$ \textcolor{white}{03}1.5  & \textcolor{white}{0}3'800 $\pm$ \textcolor{white}{0'0}70 & \textcolor{white}{0}26.4 $\pm$ \textcolor{white}{00}0.2\\
        \midrule
        \multirow{5}{*}{Zambia} & Average occupancy rate      & \textcolor{white}{-0}14.4 & 12'300 \textcolor{white}{$\pm$ 0'000} & 150.2 \textcolor{white}{$\pm$ 000.0} \\
         & RF per region \cite{stevens2015disaggregating}     & -193.4 $\pm$ 361.3 &     19'000 $\pm$ 8'000 & 271.9 $\pm$ 139.0\\ 
        & CNN                                                 & \textcolor{white}{-0}59.0 $\pm$ \textcolor{white}{00}1.6  & \textcolor{white}{0}6'600 $\pm$ \textcolor{white}{0'}300 & \textcolor{white}{0}82.7 $\pm$ \textcolor{white}{00}5.6 \\ 
        & \textsc{Pomelo}                                     & \textcolor{white}{-0}68.7 $\pm$ \textcolor{white}{00}3.1 & \textcolor{white}{0}6'100 $\pm$ \textcolor{white}{0'}200 & \textcolor{white}{0}75.6 $\pm$ \textcolor{white}{00}1.6\\ 
        \cline{2-5}
         & \textsc{Pomelo} --  7 countries         & \textcolor{white}{-0}85.4 $\pm$ \textcolor{white}{00}0.5 & \textcolor{white}{0}3'810 $\pm$ \textcolor{white}{0'0}10 & \textcolor{white}{0}49.6 $\pm$ \textcolor{white}{00}0.2\\ 
         & \textsc{Pomelo} -- Zambia only             & \textcolor{white}{-0}74.0 $\pm$ \textcolor{white}{00}1.4  &  \textcolor{white}{0}4'820 $\pm$ \textcolor{white}{0'0}70 & \textcolor{white}{0}47.6 $\pm$ \textcolor{white}{00}0.1\\
        \midrule
        \multirow{5}{*}{Mozambique} & Average occupancy rate       & \textcolor{white}{0}-88.1 \textcolor{white}{$\pm$ 000.0} & 82'500 \textcolor{white}{$\pm$ 0'000} & 153.0 \textcolor{white}{$\pm$ 000.0} \\
         & RF per region \cite{stevens2015disaggregating}                          & \textcolor{white}{0}-49.5 $\pm$ \textcolor{white}{0}21.1 & 59'000 $\pm$ 3'000 & 102.0 $\pm$ \textcolor{white}{00}3.6\\
         & CNN                                                 & \textcolor{white}{-0}43.6 $\pm$ \textcolor{white}{03}5.2  & \textcolor{white}{}35'200 $\pm$ 1'300 & \textcolor{white}{0}63.0 $\pm$ \textcolor{white}{00}1.9 \\ 
         & \textsc{Pomelo}                                      & \textcolor{white}{-0}48.3 $\pm$ \textcolor{white}{00}3.2 & 36'900 $\pm$ \textcolor{white}{0'}900 & \textcolor{white}{0}47.5 $\pm$ \textcolor{white}{00}0.6\\ 
        \cline{2-5}
         & \textsc{Pomelo} -- 7 countries                       & \textcolor{white}{-0}83.0 $\pm$ \textcolor{white}{00}0.5 & 16'800 $\pm$ \textcolor{white}{0'}200 & \textcolor{white}{0}34.1 $\pm$ \textcolor{white}{00}0.2\\ 
         & \textsc{Pomelo} -- Mozambique only                   & \textcolor{white}{-0}61.0 $\pm$ \textcolor{white}{00}3.0  & 30'600 $\pm$ \textcolor{white}{0'}900 & \textcolor{white}{0}45.0 $\pm$ \textcolor{white}{00}1.0\\
        \bottomrule
    \end{tabular}
\end{table}

\subsection*{Visual results}

\begin{figure*}[htb] %tb
    \centering
    \includegraphics[width=0.6\textwidth]{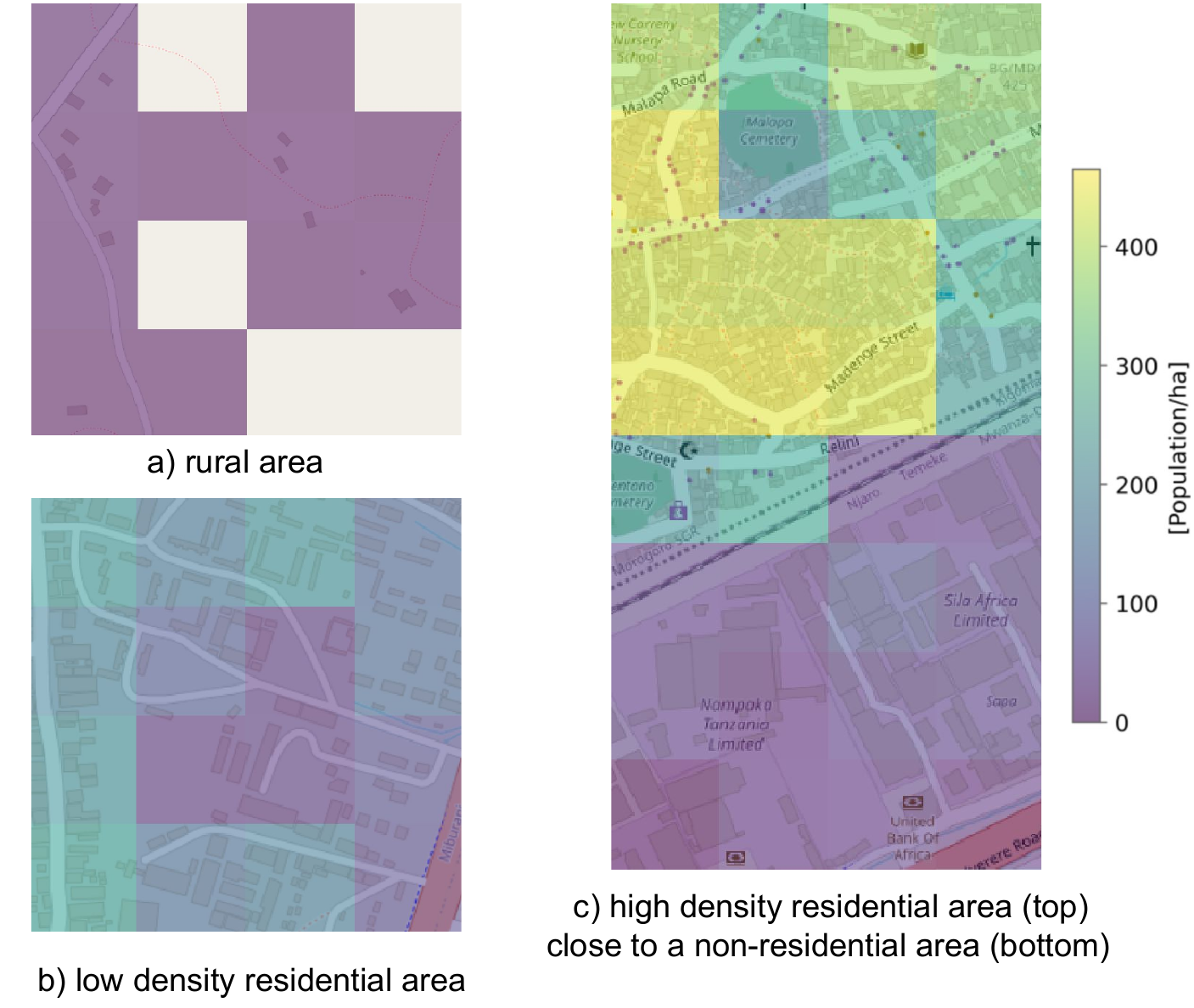}
    \caption{Visual examples of population estimates on a 100~m grid, represented as heat maps on OSM background~\cite{OpenStreetMap}. The color scale for the heat maps is shown in the top right in absolute population counts.}
    \label{fig:visual_examples}
\end{figure*}

The finest census level available for validation are administrative regions with areas of a few km\textsuperscript{2} to several hundred km\textsuperscript{2}, and ranging from very low to very high population densities.
To visually examine the predicted population maps at 100~m resolution, we select regions with known low/high densities and resort to OSM for verification. Figure~\ref{fig:visual_examples} presents example population estimates, shown as heat maps overlaid on OSM for regions around the city of Dar-es-Salaam in Tanzania. Each example covers an area of 400$\times$400~m, corresponding to 16 cells in our gridded map.
Figure~\ref{fig:visual_examples}a shows an example of a very low population estimate in a rural area.
The estimates in low-density residential areas (Figure~\ref{fig:visual_examples}b) are somewhat higher and more variable, due to varying numbers of buildings.
In high-density residential areas (top part of Figure~\ref{fig:visual_examples}c), the estimates are even higher.
%
%%The model has, however, learned to do more than just count buildings or building areas: for instance, in a non-residential area with mostly commercial buildings it predicts a relatively low and uniform density, Figure~\ref{fig:visual_examples}d.
For non-residential areas with mostly commercial buildings (bottom part of Figure~\ref{fig:visual_examples}c), the model predicts a relatively low and uniform density. 
These variations can occur over a small spatial distance: as an example, Figure~\ref{fig:visual_examples}c  show a sudden drop in population density between immediately adjacent city areas.

%When inspecting the estimated building occupancy rates, we found that \textsc{Pomelo} generally predicts higher values in larger cities than in smaller cities for residential areas with similar characteristics. Figure~\ref{fig:occupancy_rate_examples_v2}a shows a 200$\times$200~m\textsuperscript{2} area of a residential district in the moderately-sized city of Musoma (Tanzania), with an estimated (average) building occupancy rate of 2.2. In contrast, Figure~\ref{fig:occupancy_rate_examples_v2}b shows an area with a similar number of buildings in Tanzania's largest city, of Dar-es-Salaam, which has a much higher estimated occupancy rate of 3.6. 

\begin{figure*}[htb]%tb
    \centering
    \includegraphics[width=0.7\textwidth]{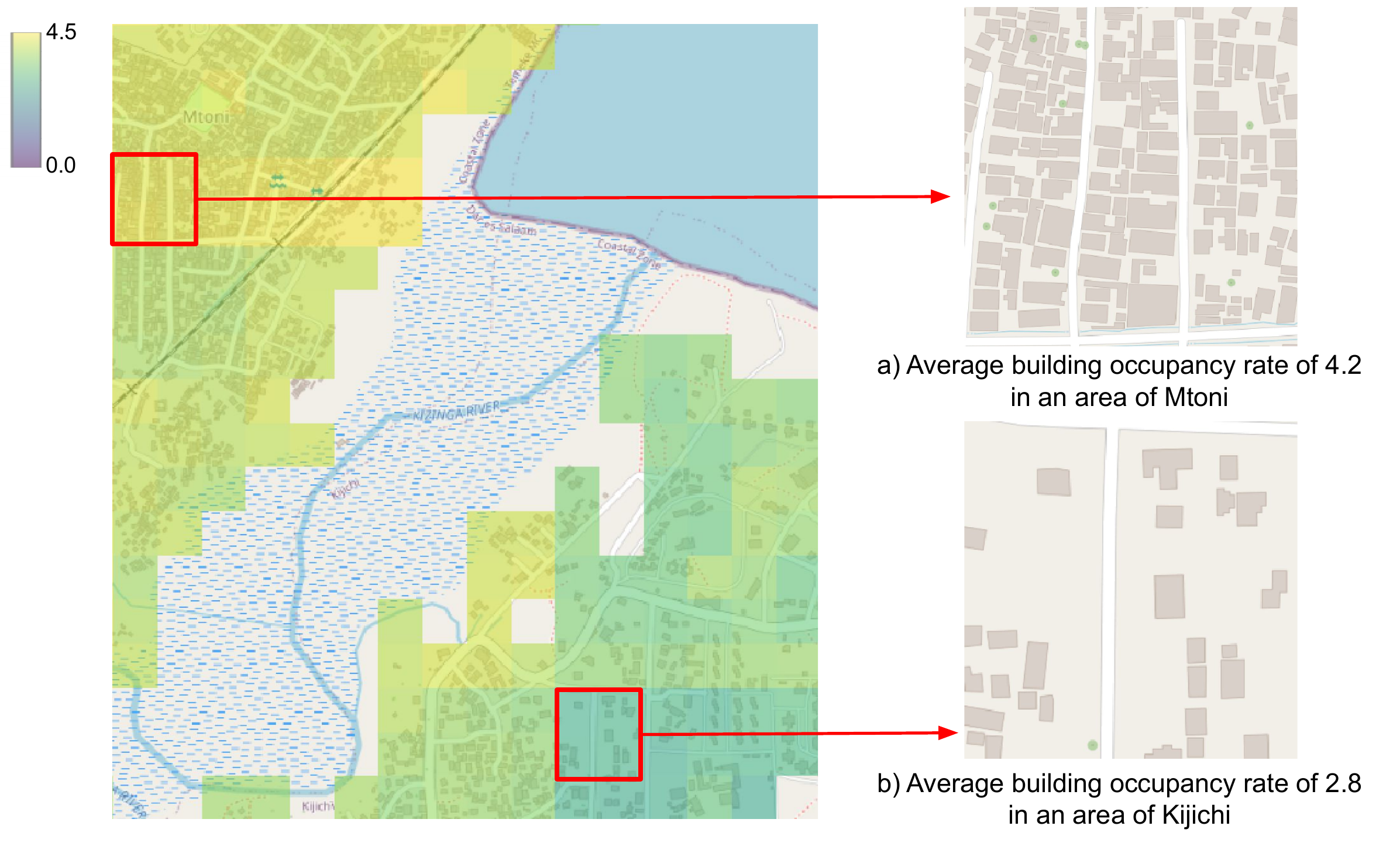}
    \caption{Estimated average building occupancy rate, depicted as heat map on OSM background, for the Mtoni and Kijichi regions of Dar-es-Salaam. Note the significantly different occupancy rates in the marked areas.}
    \label{fig:occupancy_rate_examples_v2}
\end{figure*}

Figure~\ref{fig:occupancy_rate_examples_v2} depicts the predicted building occupancy rates as heat maps for two neighborhoods in the city of Dar-es-Salaam, that are located close to each other but
feature very different occupancy rates. The densely populated area of the Mtoni region in Figure~\ref{fig:occupancy_rate_examples_v2}a has an estimated (mean) building occupancy rate of 4.2, 
whereas the area in the Kijichi region in Figure~\ref{fig:occupancy_rate_examples_v2}b has a much lower estimated occupancy rate of 2.8.

\begin{figure*}[htb] %tb
    \centering
    \includegraphics[width=0.8\textwidth]{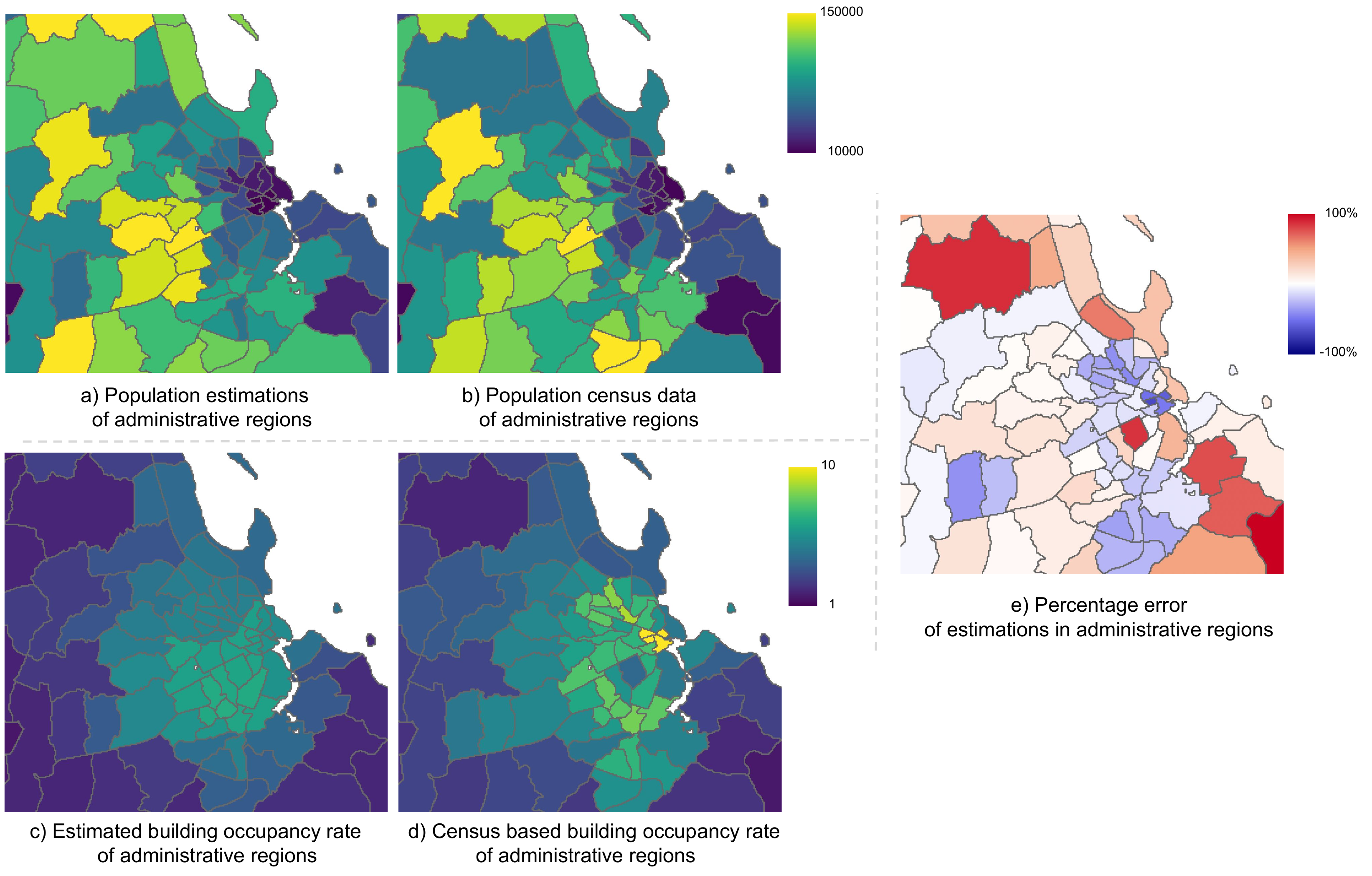}
    \caption{Population and building occupancy rate per ward around Dar-es-Salaam, Tanzania: a) estimated population counts, b) true population counts, c) estimated building occupancy rate, d) true occupancy rate, e) signed relative errors of population estimates. The map visualizations were created in QGIS 3.14 \cite{QGIS_software}}
    \label{fig:signed_error_per_region}
\end{figure*}

\begin{figure*}[htb] %tb
\centering
\begin{tabular}{ccccc} 
\includegraphics[height=4.8cm]{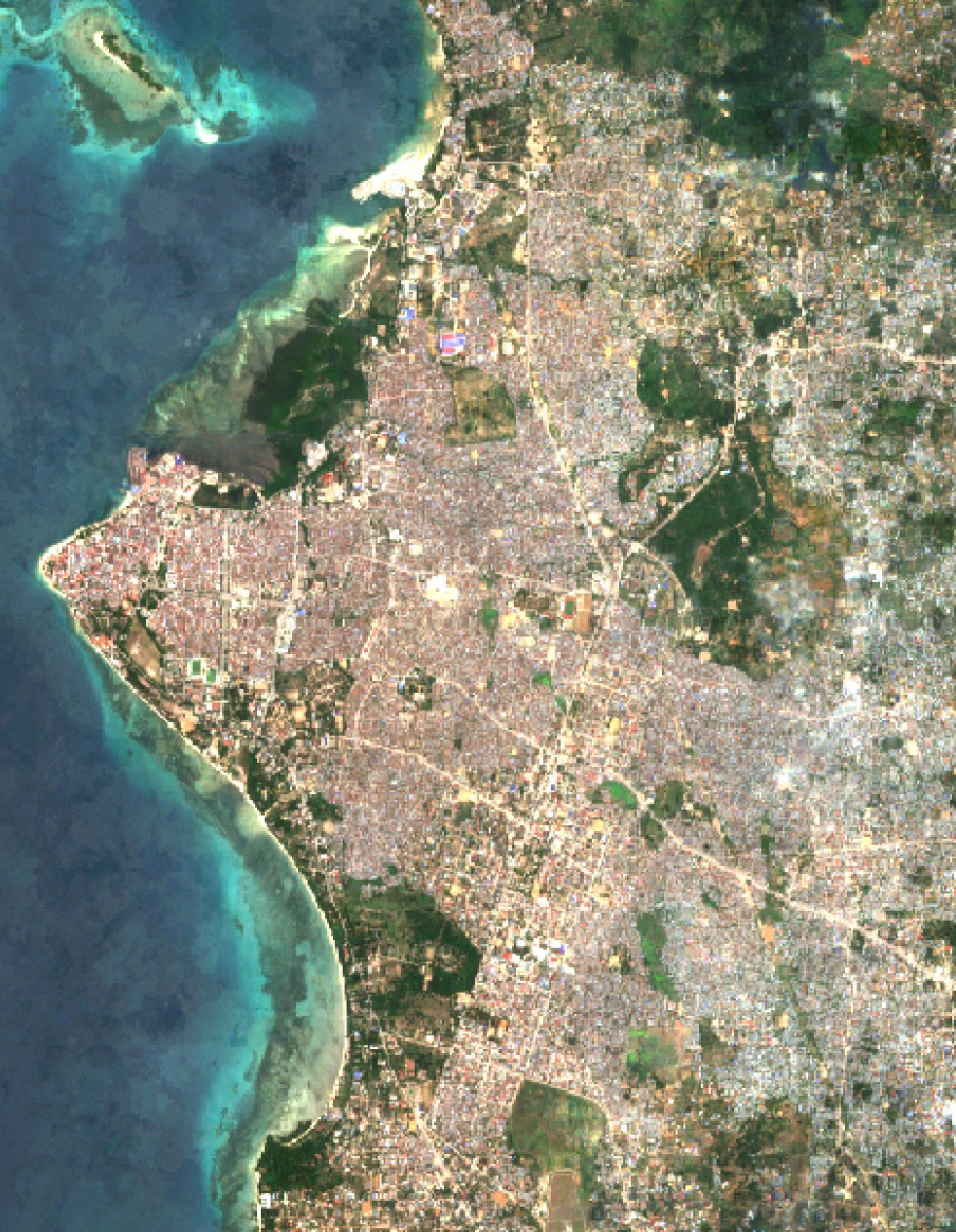} &
\includegraphics[height=4.8cm]{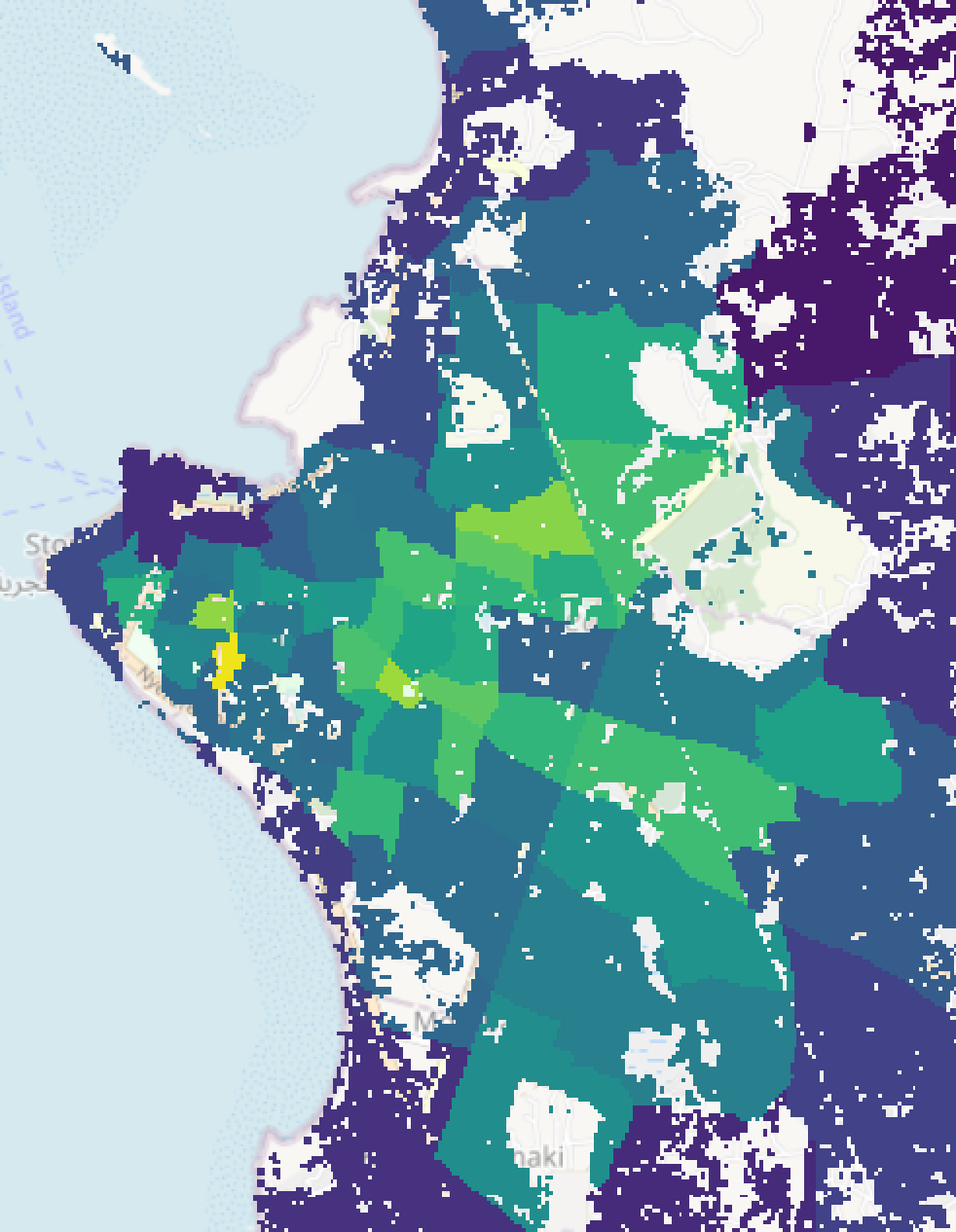} &
\includegraphics[height=4.8cm]{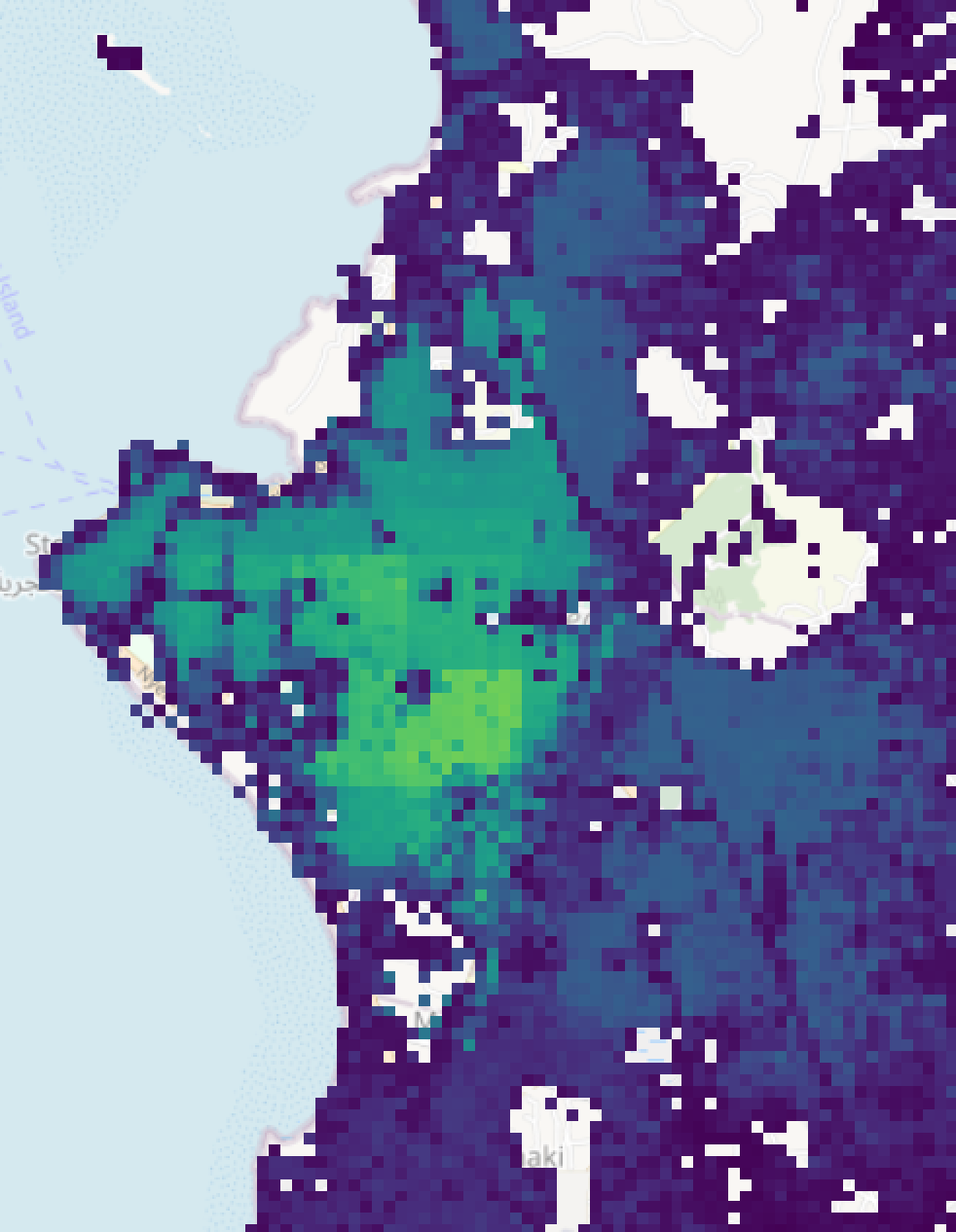} &
\includegraphics[height=4.8cm]{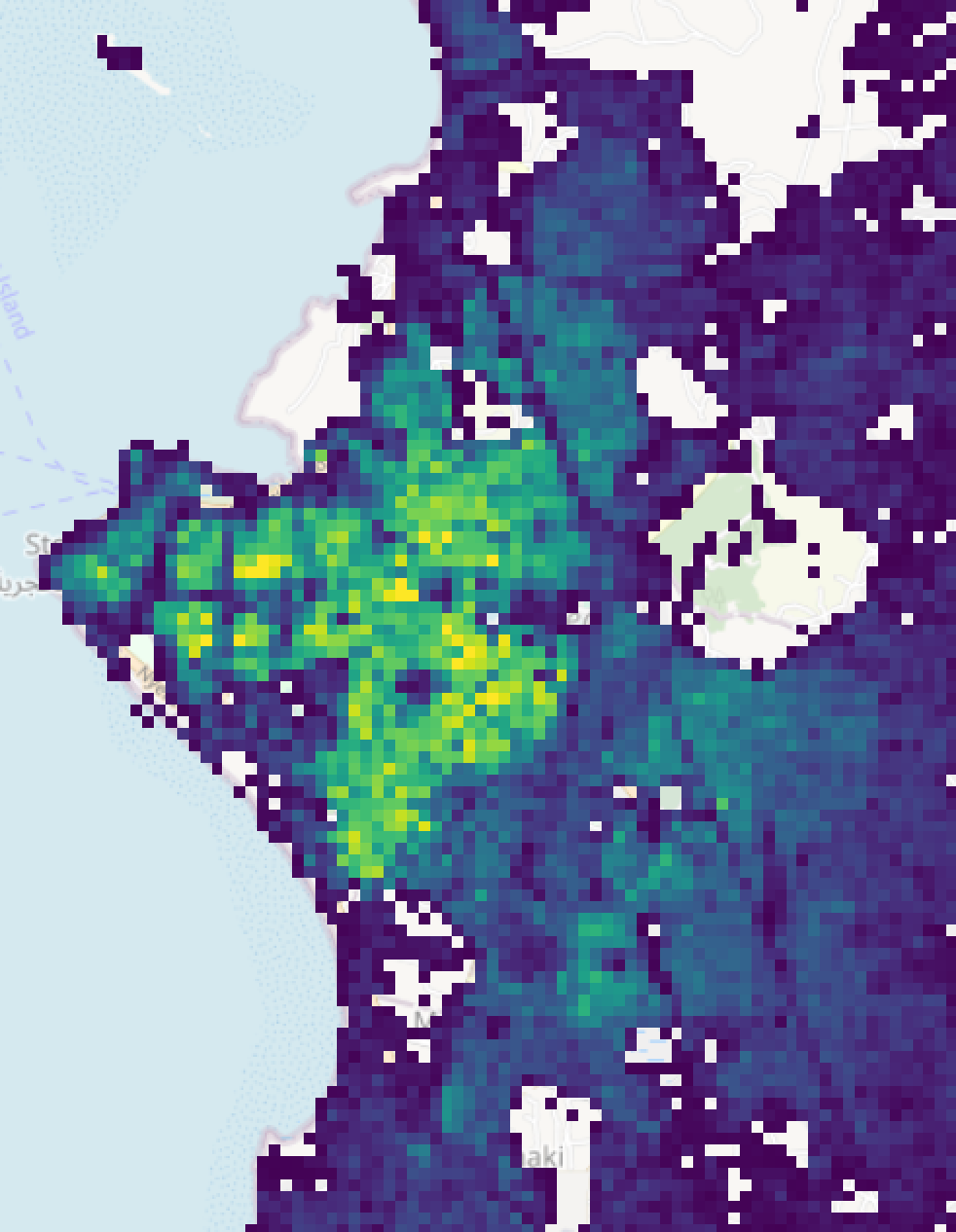} &
\includegraphics[height=4.8cm]{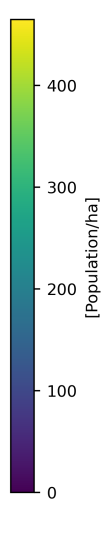} \\
a) Sentinel-2 RGB bands & b) HRPDM & c) RF per region & d) \textsc{Pomelo} &  \end{tabular} 
\caption{Visual comparison of the Sentinel-2 imagery~\cite{sentinel2data} \textit{High Resolution Population Density Maps} (HRPDM)~\cite{facebookPop} per-region RF disaggregation~\cite{stevens2015disaggregating} and \textsc{Pomelo}. Shown location: Zanzibar City, Tanzania. The Sentinel-2 mosaic was created with Google Earth Engine \cite{gorelick2017google} and visualized in QGIS~\cite{QGIS_software}.}
\label{fig:visual_baseline_comparison}
\end{figure*}

Figure~\ref{fig:signed_error_per_region} visualizes population numbers and building occupancy rates per administrative region (``ward'') in the surroundings of Dar-es-Salaam. One can see that wards have rather heterogeneous sizes and populations.
Figure~\ref{fig:signed_error_per_region}a depicts our model's population estimates, obtained by aggregating the grid cells within each ward.
Figure~\ref{fig:signed_error_per_region}b depicts the reference values from the national census. 
Figure~\ref{fig:signed_error_per_region}c are our model's estimated building occupancy rates, computed by dividing the estimated population by the number of buildings -- or, equivalently, by averaging the predicted densities, with weights proportional to the building counts in their corresponding grid cells.
Figure~\ref{fig:signed_error_per_region}d are the true occupancy rates according to the census (i.e., the population count divided by the number of buildings).
The relative error of the per-ward population estimates, as a percentage of the true numbers, is shown in Figure~\ref{fig:signed_error_per_region}e. In general, the predictions are in good agreement with the actual numbers.
%We observe the highest positive errors in built-up low-density regions, where \textsc{Pomelo} tends to over-estimate building occupancy. On the contrary, it tends to under-estimate the extremely high occupancies in the city center.
We do observe a trend that \textsc{Pomelo} overestimates the populations in built-up low-density regions, while it tends to underestimate the extremely high occupancies in the centers of large cities.
Finally, Figure~\ref{fig:visual_baseline_comparison} shows a visual comparison of recent population maps of Zanzibar City, at 100m resolution; Including the \textit{High Resolution Population Density Maps} (HRPDM) project~\cite{facebookPop}, Random Forest disaggregation per census region following~\cite{stevens2015disaggregating} (but trained with the same data as our method), and \textsc{Pomelo}. It is apparent that \textsc{Pomelo} recovers the population distribution in more detail (i.e., with a higher effective resolution), whereas the HRPDM maps as well as the region-wise disaggregation result appear overly smooth and do not recover high-frequency variations of the population density.

\section*{Discussion}
\label{sec:discusssion}

We evaluate the performance of our method using three different scenario: \emph{coarse supervision}, \emph{fine supervision}, \emph{transfer task}. 
%Besides the more common \emph{coarse supervision} scenario, we have also run our evaluations for a second setting, \emph{fine supervision}. 
%
In the past few years, the first method has been the \emph{de facto} standard for top-down population estimation~\cite{stevens2015disaggregating, grippa2019improving} and connects our evaluations to related literature. However, we argue that the \emph{fine supervision} scenario is actually more representative of the procedure one would use in practice: On the one hand, it would seem unnatural to not use the finest available spatial units if the goal is to produce the best possible population maps from a given census dataset. On the other hand, in the \emph{coarse supervision} setting one predicts population maps at a level that is already known from the census.
The coarse scenario seems to have arisen largely from the desire to evaluate pure disaggregation methods, which requires access to ground truth counts at two different levels of the spatial hierarchy.
% Additionally to these evaluation strategies, we also evaluate the models where no census counts are available

It is worth noting that the region-level RF method of Stevens \textit{et al.}~\cite{stevens2015disaggregating} does not work well if the census is only available for spatially coarse units, which is a rather frequent situation in developing countries. We see several possible reasons for this: first, there simply are fewer regions at coarser hierarchy levels and one may be left with too few training examples to learn a good model.
Second, performing feature aggregation leads to a domain shift between coarse-level training data and fine-level test data -- in particular at the pixel level.
%Second, coarser regions have larger geographic extent, thus averaging features across them becomes problematic. Finally, geographical features are not scale-invariant \textcolor{orange}{KS: reference for this?}, leading to a domain shift between coarse-level training data and fine-level test data -- in particular at the pixel level.

We found it advantageous to first estimate the spatial distribution of the building occupancy rate, and then compute population numbers by multiplying the occupancy rate and the building count at a given location. This is in contrast to methods that directly estimate absolute population numbers~\cite{stevens2015disaggregating} or relative population fractions (which are the weights for dasymetric disaggregation~\cite{gervasoni2018convolutional}).
The finding that factoring into building counts and occupancy reduces the estimation errors lends support to two assumptions implicit in our approach: First, the available maps of building counts (respectively, building footprints) are apparently rather accurate, so that using them directly rather than as one of many ``soft'' covariates reduces the estimation error. Second, the occupancy rate appears to have lower spatial variability than the population density, making it easier to estimate from the same covariates.

The important role of country-scale building counts (respectively, building footprint maps) also merits some discussion. For our work we obtain them from free sources, namely the Open Buildings dataset~\cite{sirko2021continental} and the Grid Maps of Building Patterns~\cite{dooley2020gridded}.
Both these datasets are created with computer vision-based building detectors on the basis of high-resolution satellite imagery (GSD \textless1$\,$m).
Although the datasets are of high quality, such large-scale maps inevitably contain errors and data gaps, especially in rural areas \cite{vargas2019correcting}. We have tried to maximize completeness by fusing the two building datasets, but note that missing buildings (and to a lesser degree perhaps also spurious buildings) may cause errors in our population maps, mainly in scarcely populated areas.

Although the mentioned building footprint datasets are available for free, the underlying high-resolution images used to produce them are not. The processing of high-resolution data at country scale also requires considerable computational resources. Our method thus critically depends on data that are oftentimes primarily produced for a different purpose or for philanthropic reasons, and over whose production we have no control.
If at some point in the future no up-to-date building dataset is at hand, one could try to resort to autoregressive settlement growth models~\cite{nieves2020annually, nieves2020predicting}.
Alternatively, one may keep the building detection implicit and also feed high-resolution images to the population estimator. Such approaches have been developed~\cite{weber2018census,neal2022census}, but since they rely on (commercial) high-resolution satellite imagery they can hardly be scaled up to entire countries, except by entities who could equally supply the building counts.
Crowdsourced data (e.g., OSM roads) have their own challenges due to incompleteness and temporal inaccuracies. However, in several regions of developing countries, commercial products and official data are not available, making crowdsourced data the only source of information with sufficient coverage in those regions~\cite{barrington2017world}. 
%Some projects like the Humanitarian OSM Team are making progress in mapping those regions. 

\begin{figure} [htb]
\textcolor{blue}{
    \centering
    \includegraphics[width=0.5\textwidth]{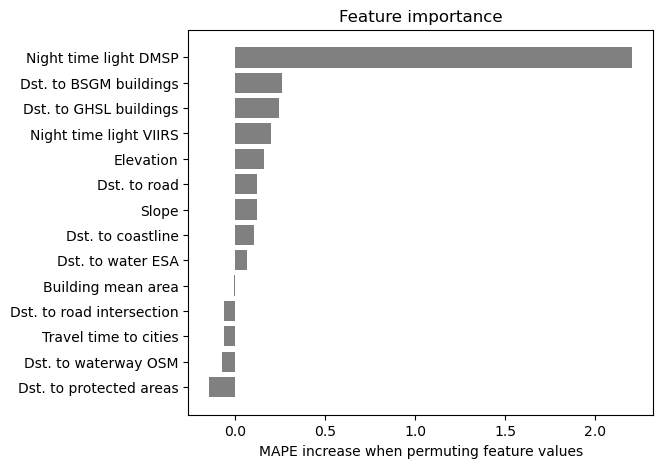}
    \caption{Feature importance analysis for the fine supervision setting in Tanzania.}
    \label{fig:feat_importance}
}
\end{figure}

In addition to the building counts, \textsc{Pomelo} is driven by several other geospatial data layers that are publicly available and correlate with population.
We have analyzed feature importance with the permutation method~\cite{breiman2001random}, which essentially measures the performance drop caused by randomly shuffling a single input layer, so as to render that covariate uninformative.
Exemplary feature importance scores for Tanzania (fine supervision) are shown in Figure~\ref{fig:feat_importance}.
%
%It turns out that the most predictive input (besides the building counts) are night light images
Although the ordering of the covariates fluctuates depending on the chosen performance measure, we noted one commonality, namely the nightlight and settlement layers are among the most predictive inputs.
%although the relative order depends somewhat on whether one optimizes for absolute error, relative error, or $R^2$.
Moreover, also features with low scores appear to carry some relevant information, as explicit feature selection based on the importance scores tends to harm model performance. For more details please refer to the Methods section.

%\textcolor{blue}{Note that for several features their importance scores are below the $\pm$ 0.2 percent points standard deviation between training runs (see Table~\ref{tab:fine_task_all}).} 

Apart from the covariates used in the present work, there are others that could potentially be useful, and that one could obtain from open geospatial data sources. For instance, online land use maps as available in OSM could help to identify non-residential buildings such as schools or shopping centers,
and building height estimates from Esch \textit{et al.}~\cite{esch2022world}, once publicly available, could help to predict more precise building occupancy rates.
Furthermore, data extracted from social media could possibly also support population mapping in certain areas. For example, the Facebook Marketing API \cite{FacebookMarketingAPI} allows one to access (anonymized and aggregated) information about the platform's users at a certain location~\cite{fatehkia2020relative,leasure2022ukraine}, such as the number of active users, the type of internet connection used, etc. It is quite possible that such information is to some degree predictive of population density. 
However, social media and OSM data are typically incomplete and biased, which is a challenge and would require computational methods that can exploit the important features without relying on their completeness.
%
%\textcolor{blue}{
% Given that there are several potential data sources that we could include to improve the model performance it would also be interesting to research feature selection methods for neural network models to identify the most relevant ones.
%Furthermore, an exploration of additional data sources and more sophisticated feature selection methods may also lead to more sound insights about the topic of population mapping.
%}
%
% In future work, we plan to extend the method to handle inhomogeneous and varying sets of inputs.
Finally, it would be important to develop techniques to handle the case where covariates are missing for certain regions or time periods in a given country.

\section*{Methods}
\label{sec:methods}

\subsection*{Data}
\label{data}

To validate our proposed methodology, we use covariates that are related to population, and that can be derived from remote sensing imagery, open geo-spatial data (e.g., OSM) or governmental sources. For the present study we rely on data that have been preprocessed by the WorldPop~\cite{worlpop2022} project. 
We collect the covariates listed in Table~\ref{tab:covars}, with a resolution of 100 $\times$ 100~m. A selection of variables is visualized in the left part of Figure~\ref{fig:methodology}. Moreover, we obtain census data for the countries of Tanzania (ward-level, $n=3654$) Zambia (ward-level, $n=1421$)
and Mozambique (\textit{postos administrativos}-level, $n=413$) to evaluate the performance of our proposed \textsc{Pomelo} method.  

% \usepackage{booktabs}
% \usepackage{multirow}
\begin{table}[htb]
%\scriptsize
\caption{Summary of the used covariates.}
\label{tab:covars}
\begin{tabularx}{\textwidth}{lrX}
\toprule
Type            & ~ & Description                       \\ \midrule
\multirow{4}{*}{Buildings}
            & 1 & Building counts from Google Open Buildings~\cite{sirko2021continental}               \\ %\arrayrulecolor{gray}\midrule
            & 2 & Building counts from Gridded Maps of Building Patterns~\cite{dooley2020gridded}   \\ %\arrayrulecolor{gray}\midrule
            & 3 & Mean building areas from Google Open Buildings~\cite{sirko2021continental}              \\ %\arrayrulecolor{gray}\midrule
            & 4 & Mean building areas from Gridded Maps of Building Patterns~\cite{dooley2020gridded}   \\ \arrayrulecolor{black}\midrule
Accessibility   & 5 & Travel time to city with more than 50k inhabitants~\cite{worlpop2022}    \\ \arrayrulecolor{black}\midrule
\multirow{2}{*}{Nightlight}
            & 6 & Cloud-free DMSP nightlight composite~\cite{elvidge2013viirs}                                  \\ %\arrayrulecolor{gray}\midrule
            & 7 & Cloud-free VIIRS nightlight composite~\cite{elvidge2013viirs}                               \\ \arrayrulecolor{black}\midrule
\multirow{2}{*}{Settlement}
            & 8 & Distance to built-up area from Global Human Settlement layer (GHSL)~\cite{GHSL2018}     \\ %\arrayrulecolor{gray}\midrule
            & 9 & Distance to built-up area from Built-Settlement Growth Model (BSGM)~\cite{worlpop2022}   \\ \arrayrulecolor{black}\midrule
\multirow{2}{*}{Topography}       
            & 10 & Elevation model from the Shuttle Radar Topography Mission (SRTM)~\cite{farr2007shuttle}     \\ %\arrayrulecolor{gray}\midrule
            & 11 & Slope from the Shuttle Radar Topography Mission (SRTM)~\cite{farr2007shuttle}                \\ \arrayrulecolor{black}\midrule
\multirow{3}{*}{Water}          
            & 12 & Distance to Waterbody from ESA CCI Water~\cite{lamarche2017compilation} \\ %\arrayrulecolor{gray}\midrule
            & 13 & Distance to Waterway from OpenStreetMap~\cite{OpenStreetMap}      \\ %\arrayrulecolor{gray}\midrule
            & 14 & Distance to Coastline~\cite{worlpop2022}           \\ \arrayrulecolor{black}\midrule
Land cover  & 15 & Distance to Nature Reserves from World Database on Protected Areas (WDPA) \cite{WDAP2021}   \\ \arrayrulecolor{black}\midrule
\multirow{2}{*}{Roads}
            & 16 & Distance to Road from OpenStreetMap~\cite{OpenStreetMap}      \\ %\arrayrulecolor{gray}\midrule
            & 17 & Distance to Road Intersection derived from OpenStreetMap~\cite{OpenStreetMap}      \\ \arrayrulecolor{black}\bottomrule
\end{tabularx}
\end{table}

We also aggregate the census counts to the second-finest administrative level for the countries of Tanzania (district-level), Zambia (constituency-level), and Mozambique (district-level), using administrative boundaries from the Humanitarian Data Exchange~\cite{humdata2022}, to simulate scenarios where the census counts are coarser. We will later refer to these aggregated data as \textit{coarse census} counts.
%, with administrative region counts of 170 and 156 for Tanzania and Mozambique, respectively.
Finally, to assess the generalization performance of \textsc{Pomelo} across countries, we collect the same source data for four further countries, Uganda (UGA), Rwanda (RWA), Nigeria (NGA), and Democratic Republic of Congo (COD), to be used as additional training data.
The dataset is summarized in Table~\ref{tab:census_sizes}.

% \usepackage{booktabs}
\begin{table}[tb]
\centering
\small
\caption{Overview of census data used and data extent in \# of pixels. Each pixel covers a region of 100 $\times$ 100~m. The Category \textit{Others} refers to the additional countries of Uganda, Rwanda, Nigeria and the Democratic Republic of Congo.%\bk{}{, used exclusively to evaluate generalization performance of the model} \nando{}{, used exclusively for training when tasks}.
}
\begin{tabular}{@{}l|cc|cc|cc@{}}
\toprule
&\multicolumn{2}{c|}{\# census regions} & \multicolumn{2}{c}{\# pixels} & \multicolumn{2}{|c}{Avg. fine census region} \\
& fine   & coarse  & total & built-up & area & population\\ \midrule
Tanzania      & 3654 & 170 & 111M & 7.5\%                                        &  \textcolor{white}{0'}333~km$^2$  &  \textcolor{white}{0}15'449 \\
Zambia        & 1421 & 150 & \textcolor{white}{0}90M      & 3.4\%                &  \textcolor{white}{0'}633~km$^2$  &  \textcolor{white}{0}13'035 \\
Mozambique    & \textcolor{white}{0}413 & 156 & \textcolor{white}{0}96M & 5.6\%  &  2'324~km$^2$                     &  \textcolor{white}{0}75'095    \\
Others        & 2779 & 295 & 412M & 5.5\%                                        &  9'975~km$^2$                     &   139'171       \\ \bottomrule
\end{tabular}
\label{tab:census_sizes}
\end{table}

\subsection*{Proposed method: Population Mapping by Estimating Local Occupancy Rates}
\label{pomelo}

%\textcolor{orange}{KS: Should we switch the order to the one employed in geo-science journals, with the "Materials and Methods" section at the end? Just a thought.\\}
%\devis{}{It's is quite uncommon in RSE. If you want to submit to something like PNAS ok, but then the paper must be really less tech-centric. I will not do it}

Our goal is to produce population maps with a finer resolution than the underlying census counts. In machine learning terms we are thus faced with an instance of \emph{weakly supervised} learning~\cite{zhou2018brief}:
we do not have access to ground truth values for the individual grid cells, rather we only have one target value (the census count for a region) as supervision signal for a whole set of outputs (all grid cells within the region).

\begin{figure*}[ht]
    \centering
    \includegraphics[width=\textwidth]{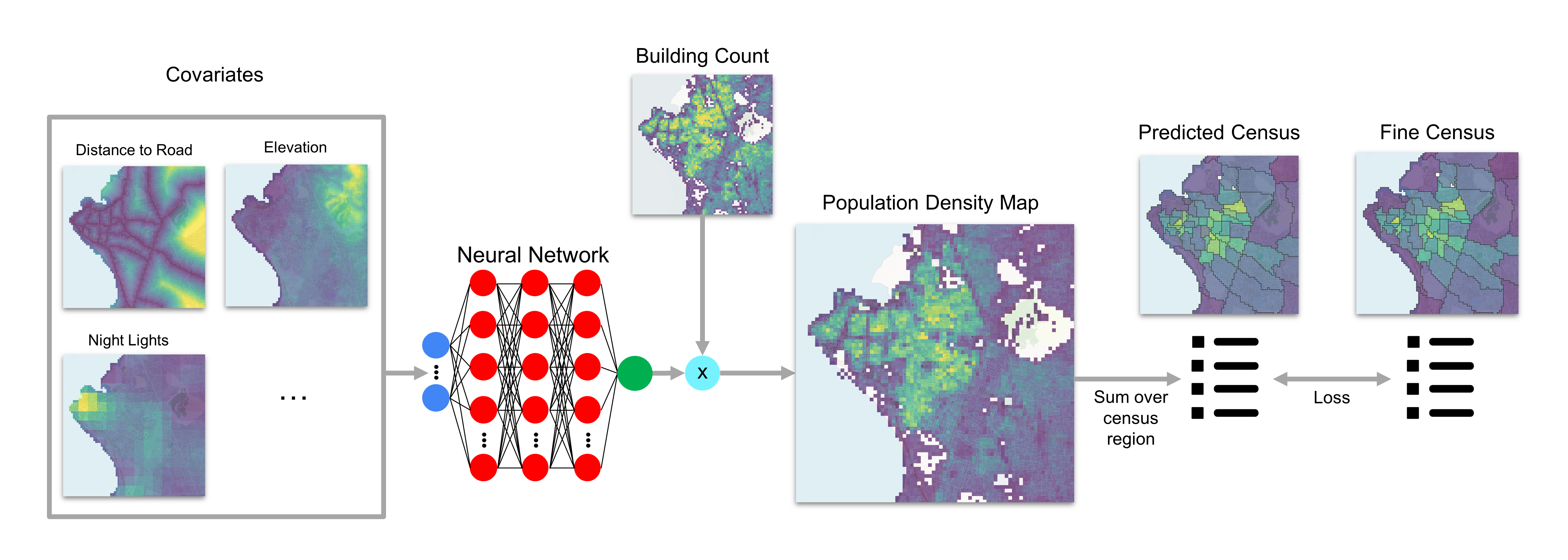}
    \caption{Overview of the proposed method: We use a neural network to create a mapping between covariates (blue circles) and population density. The direct output of the neural network are building occupancy rates (green circle) which we multiply with the building count map to obtain population counts. We receive the supervision signal for training by aggregating the population maps to the respective census regions through summation, and comparing our estimates with the available census data.}
    \label{fig:methodology}
\end{figure*}

This cumulative sum per region is the supervision signal used by our model. More specifically, we employ a neural network, c.f.~Figure~\ref{fig:methodology}, that maps the covariates to population density estimates and that can be trained with lower-resolution ground truth: rather than per-pixel supervision, which is not available, the training procedure compares \emph{aggregated} estimates per administrative region to the available region-level census data.

We denote the covariates at grid location $l_i$ as $X(l_{i})$, the population estimate at the same location as $\hat{p}\left(l_{i}\right)$, and $c_{j}$ as the true census data for region $A_{j}$.
We observe that the correlation between the number of buildings at a given location and its population turns out to be so strong that even dasymetric disaggregation based \emph{only} on building counts produces reasonable (albeit less precise) population density maps, as shown in the Results section.
We exploit this strong correlation for the design of the neural network's architecture as follows:
To obtain even more accurate estimates, 
%this relation can be explicitly modeled by setting up the neural network to predict 
we predict a spatially varying \emph{building~occupancy~rate}~$f_{\theta}(X(l_{i}))$, rather than the population count directly. The population estimate $\hat{p}\left(l_{i}\right)$ is then computed from it as
\begin{equation}
    \hat{p}\left(l_{i}\right) = \mathit{f_{\theta}}\big(X(l_{i})\big) \cdot b\left(l_{i}\right),
\end{equation}
where $b\left(l_{i}\right)$ is the number of buildings at location $l_{i}$ (obtained from existing building products~\cite{sirko2021continental,dooley2020gridded}) and $\theta$ are the parameters of the neural network. 
We use a neural network composed of a sequence of convolutional layers with kernel sizes $1 \times 1$, i.e., its architecture is equivalent to a per-pixel multi-layer perceptron (MLP).
Empirically, we observe that adding more spatial context (e.g., via convolutional layers with kernel sizes \textgreater 1) does not significantly improve the model's prediction performance (see results in Table~\ref{tab:kernels}). At first this may seem surprising, but at the low GSD of our maps most covariates are smooth. Consequently, the disadvantages of a large receptive field outweigh the potential benefits: feature values of neighboring pixels tend to be very similar and hence do not add much information. Moreover, the spatial averaging effect still causes some loss of high-frequency detail (a similar effect has been observed by other authors, e.g., de Lutio \textit{et al.}~\cite{Lutio_2019_ICCV}). Also, the increased number of parameters is more prone to overfitting.
A practical benefit of using 1$\times$1 kernels is that, in the absence of spatial interactions, there is no spatial diffusion of the input information. All computations can be restricted to the small fraction of grid cells with at least one building. This makes the network memory-efficient, an important feature when processing large countries with coarse census regions, as GPU memory is often the computational bottleneck for modern neural networks.
% TODO:
% - Maybe add the description of the head and loss function, something similar to the poster

% Loss function
%Our goal is to produce population maps with a finer resolution than the underlying census counts. In machine learning terms we are thus faced with an instance of \emph{weakly supervised} learning:
%we do not have access to ground truth values for the individual grid cells, rather we only have one target value (the census count for a region) as supervision signal for a whole set of outputs (all grid cells within the region).
%
The fundamental insight behind \textsc{Pomelo} (and any other top-down disaggregation scheme) is that, given enough data, the weak supervision is sufficient:
the sum over many grid cells is only a weak constraint per census region, but it accumulates over different regions because similar covariate features at different locations must yield similar population densities; whereas the input covariates have (at least) the target resolution and inject the missing high-frequency information.
Formally, measuring the loss $\mathcal{L}$ over all grid cells $l_i$ in each region $A_j$ gives rise to the following optimization problem:
% We view the problem as a weakly supervised objective function such that it enables us to take advantage of the raw, pixel-wise features and the population census data, which is only available at the level of administrative regions. This gap in resolution between the coarse scale training data (administrative regions) and fine scale predictions (regular grid smaller than administrative regions) is what characterizes this formulation as a weakly supervised problem. We aggregate the pixel-wise predictions to their respective administrative regions to obtain the predicted census, such that we can formulate the optimization problem as:
%

\begin{equation}
\label{eq:superres}
\begin{split}
&\hat{\theta}=\underset{\theta}{\operatorname{argmin}} \sum_{j} \mathcal{L}\left(\hat{c}_{j}, c_{j} \right)\;,\\
&\text{with } \hat{c}_{j} = \sum_{i \in A_{j}} \hat{p}\left(l_{i}\right).
\end{split}
\end{equation}
Empirically, the loss function that achieves the best results is the $L_1$ distance between the log-transformed predictions and targets:
\begin{equation}
    \mathcal{L}\left(\hat{c}_{j},c_{j}\right) =  | \log \hat{c}_{j} - \log c_{j}\,|\;.
\end{equation}

If census data are available at inference time, we can leverage the associated, additional constraints to perform disaggregation (dasymetric mapping) as a form of post-processing. 
In other words, we treat the final outputs of our model $\hat{p}\left(l_{i}\right)$ not as absolute quantities but as relative proportions. By linearly rescaling the predictions within a census region $j$ such that they add up to 1, we obtain weights that indicate what fraction of the total census count $c_j$ should be assigned to each location $l_i$. By multiplying those weights with the $c_j$, we then obtain the disaggregated population estimates $\hat{p}_{adj}$, adjusted to exactly match the coarse census counts:
\begin{equation}
\hat{p}_{adj}\left(l_{i}\right)= \frac{\hat{p}\left(l_{i}\right)}{\sum_{k \in A_{j}} \hat{p}\left(l_{k}\right) } \cdot c_{j}.
\end{equation}
As long as the census data are correct, that adjustment can be expected to improve the population density map compared to the raw, absolute estimates. If no census counts are available at all, we just keep the raw estimates $\hat{p}\left(l_{i}\right)$.

\subsection*{Model Setup}
\label{sec:model_setup}

Our neural network is composed of four convolutional layers with kernel sizes $1\times1$ or, equivalently, of a four-layer MLP that is applied per pixel. The first three layers each have 128 filters, and the last layer has one filter to output a scalar density value per location. We use dropout~\cite{srivastava2014dropout} (with probability 0.4), and rectified linear unit (\textit{ReLU}) layers after each convolutional layer.
We apply a \textit{softplus} function to the output of the last convolutional layer, constraining the occupancy rates to positive numbers. The model is trained using the Adam optimizer~\cite{kingma2014adam} with a base learning rate of $0.0001$.
%, and weight decay of $0.01$, for regularization. 
The weight decay parameter for regularization is optimized via grid-search on the validation set.
To mitigate the low number of training samples, we propose a data augmentation strategy specific to our task, namely we create artificial "pseudo-regions" from two real administrative regions, by merging their pixels and summing their population counts.

The importance of the main components of our model design is ablated in Table ~\ref{tab:ablation_tza}. 
%Finally, we perform systematic experiments to empirically support the proposed design.
%Table~\ref{tab:ablation_tza} 
It shows quantitative results obtained with different model setups for the scenario with \textit{coarse supervision in Tanzania}.
We start from a baseline that has the same model architecture, but
\begin{enumerate}
    \item employs the standard $L_1$ distance between predictions and target values as loss function (no $\log L_1$ loss);
    \item directly predicts population counts per grid cell, and
    \item is trained only with the actual administrative regions (no region-based augmentation).
\end{enumerate}
We then gradually add the components used in our proposed setting. First, we find that computing the loss with log-transformed outputs significantly improves over the standard $L_1$ loss. 
%(and also the $L_2$ loss, not shown).
%
Second, predicting the building occupancy rate instead of the population and transforming it to a population count in postprocessing also clearly reduces the error, especially the $R^2$ metric.
Third, data augmentation by synthetically merging census regions also has a positive effect, confirming the beneficial effect of problem-specific augmentation when training data is scarce.
Finally, combining all three measures, as in our proposed model, yields greatly improved predictive skill, reaching $\approx$25\% lower mean absolute error, respectively \textgreater14 percent points higher $R^2$.

\begin{table}[tb]
\caption{Effect of modeling choices (Tanzania data, coarse supervision).}
\label{tab:ablation_tza}
\centering
%\footnotesize
\begin{tabular}{@{}lcccccc@{}}
\toprule
Method &
LogL1 &
Occ. &
Aug.  &
$R^2$ $\uparrow$ [\%] &
MAE $\downarrow$ &
MAPE $\downarrow$ [\%] \\ \midrule
Baseline   &   &   &    & 71.4 $\pm$ 0.3 & 4'130 $\pm$ \textcolor{white}{1}80 & 27.5 $\pm$ 0.9 \\
$\,\,$+ Log$L_1$ loss  & \checkmark &   &    & 73.2 $\pm$ 0.3 & 3'580 $\pm$ \textcolor{white}{1}30 & 23.0 $\pm$ 0.2 \\
$\,\,$+ Occupancy rate  &   & \checkmark &    & 81.4 $\pm$ 1.2 & 3'820 $\pm$ 180 & 26.8 $\pm$ 1.2 \\
$\,\,$+ Augmentation  &   &   & \checkmark & 73.0 $\pm$ 1.2 & 3'820 $\pm$ \textcolor{white}{1}30 & 25.0 $\pm$ 0.4  \\
%\rowfont{\color{Tomato}} + LogL1 + Occ & x & x &  & 23.4$\pm$0.4 & 84.3$\pm$1.0 & 3'330$\pm$60 \\
\textsc{Pomelo}       & \checkmark & \checkmark & \checkmark  & 85.8 $\pm$ 0.7 & 3'100 $\pm$ \textcolor{white}{1}30 & 21.7 $\pm$ 0.3 \\
\bottomrule
\end{tabular}
\end{table}

%%%QQQ

    Table~\ref{tab:kernels} show the effect of larger kernel widths on \textsc{Pomelo}'s performance for Tanzania, in both the coarse and fine supervision settings. It can be seen that larger kernel sizes are consistently detrimental, it appears that
    the they increase model complexity beyond the level that can be learned with the available supervision.
    On the one hand the bigger kernels inflate the number of weights that must be trained ($9\times$ more, respectively $25\times$ more).
    These additional degrees of freedom increase the risk of overfitting and make learning harder -- particularly in our weakly supervised setting, which becomes apparent with coarse supervision from only 170 regions.
    While on the other hand the covariate maps in general exhibit a high degree of spatial smoothness and do thus do not contain high-frequency context that larger kernels could capture.

    \begin{table}[!htbp]
    
        \centering
        \caption{Performance with varying kernel sizes on the dataset of Tanzania.}
        \label{tab:kernels}
        %\scriptsize
        \begin{tabular}{@{}llllll@{}}
            \toprule
            Training Setting & Kernel Size   & \#params &  \multicolumn{1}{c}{$R^2 \uparrow$ [\%]}  & \multicolumn{1}{c}{MAE $\downarrow$} & \multicolumn{1}{c}{$\!\!$MAPE $\downarrow$ [\%]}
            \\ \midrule 
            \multirow{3}{*}{coarse} 
                & $1 \times 1$ (ours)    & \textcolor{white}{0}34k  & 85.7 $\pm$ 0.9  & 3'100 $\pm$ \textcolor{white}{0}40  & 21.6 $\pm$ 0.2 \\
                & $3 \times 3$           & 313k & 83.5 $\pm$ 1.3  & 3'700 $\pm$ 110                     & 26.6 $\pm$ 0.7 \\   
                & $5 \times 5$           & 870k & 80.9 $\pm$ 1.0  & 3'900 $\pm$ 100                     & 28.4 $\pm$ 0.8 \\ 
            \midrule
            \multirow{3}{*}{fine}
                & $1 \times 1$ (ours)        & \textcolor{white}{0}34k & 87.6 $\pm$ 0.2 & 2'890 $\pm$ \textcolor{white}{0}20 & 20.4  $\pm$ 0.3 \\
                & $3 \times 3$               & 313k & 86.0 $\pm$ 2.1 & 3'010 $\pm$ 100                    & 20.9  $\pm$ 0.5 \\ 
                & $5 \times 5$               & 870k & 86.8 $\pm$ 1.0 & 3'030 $\pm$ \textcolor{white}{0}50 & 21.5  $\pm$ 0.4 \\ 
            \bottomrule
        
        \end{tabular}
    \end{table}

    %Finally, we performed an experiment to determine the top 5 most important covariates on the validation set via the permutation method (see Figure~\ref{fig:feat_importance}). 
    %We then retrain the model with the reduced set of covariates and report the numbers in Table~\ref{tab:POMELOtop5}.
    We also empirically study the choice of input covariates. To that end, we only keep the five input layers with the highest individual importance according to the permutation method~\cite{breiman2001random} and discard the remaining features. The results with this reduced input set for Tanzania (fine supervision setting) are reported in Table~\ref{tab:POMELOtop5}. Empirically, explicit variable selection noticeably harms performance. Apparently features with lower importance scores still carry valuable information that complements the most important features. In this context, note that in the proposed scheme the number of input layers is not critical: with the \textsc{Pomelo} architecture, an additional covariate only introduces 128 learnable parameters. The computational savings achievable with feature selection are negligible.
 
    \begin{table}[!h]
        \centering 
        \caption{Performance on Tanzania data with reduced covariate set (fine supervision setting).}
        \label{tab:POMELOtop5}
        %\scriptsize
        \begin{tabular}{@{}llllll@{}}
            \toprule
            Method   &  \multicolumn{1}{c}{$R^2 \uparrow$ [\%]}  & \multicolumn{1}{c}{MAE $\downarrow$} & \multicolumn{1}{c}{$\!\!$MAPE $\downarrow$ [\%]}
            \\ \midrule 
                all features (ours) & 87.6 $\pm$ 0.2 & 2'890 $\pm$ 20 & 20.4  $\pm$ 0.3 \\
                top 5 features      & 85.7 $\pm$ 0.4 & 2'980 $\pm$ 10 &  20.7 $\pm$ 0.0 \\ 
            %\midrule
            %\multirow{2}{*}{Zambia} 
            %    & all features (ours) & 88.7 $\pm$ 0.3 & 3'650 $\pm$ 20 & 48.0 $\pm$ 0.6 \\
            %    & top 5 features      & 87.9 $\pm$ 0.3 & 3'750 $\pm$ 40 & 50.1 $\pm$ 0.6 \\ 
            \bottomrule
        \end{tabular}
    \end{table}

%\subsection*{Baseline methods} 
\subsection*{Methods used for numerical comparisons} 

We compare \textsc{Pomelo} with four other methods:

\vspace{1em}
\noindent
\underline{\itshape Building count disaggregation:} Dasymetric disaggregation using only the available building counts per pixel as weights.

\vspace{1em}
\noindent
\underline{\itshape Random Forest (RF) at region level:}
Our own re-implementation of the random forest model used by WorldPop~\cite{stevens2015disaggregating}. The model is fed the same covariates (features) as the \textsc{Pomelo} network. The training units for that scheme are not grid cells but administrative regions, with features aggregated over all pixels within a region. The map visualizations were created in QGIS 3.14 \cite{QGIS_software}.

\vspace{1em}
\noindent
\underline{\itshape Markov Random Field (MRF):}
Here, disaggregation is based directly on the assumption that locations with similar features have similar population densities.
To enforce such prior knowledge explicitly, we resort to an MRF~\cite{besag86} that compares pixel-wise predictions to each other and to their encompassing region. This involves finding the combination of population values that minimizes the following (negative log-likelihood) energy function:

\begin{align}
\mathbf{\hat{p}}_{MRF} = \argmin_{\mathbf{p} \in \mathbb{R}^N}  \sum_{i}^{N} \sum_{k \in Q_{i}} | p_i - p_k | + \lambda \sum_{j} | c_j - \sum_{k \in A_j} p_k |\;,
\end{align}
where $N$ is the number of grid cells and $Q_i$ the set of nearest neighbors to cell $l_i$ in the (normalized) feature space. The first term encourages locations with similar features (i.e., the $k$ nearest neighbors to pixel $i$ in feature space) to have similar population counts; the second term pushes the aggregate count over an administrative region towards the region's total census. Parameter $\lambda$ determines the balance between the two constraints; for all the experiments we set $\lambda = 1$.
We create the (approximate) nearest-neighbor graph with the fast ANN method~\cite{johnson2019billion}, initialize the population counts with the \textit{building count disaggregation} described above, and find an optimal configuration by minimizing the energy function with the Iterated Conditional Modes algorithm~\cite{besag86}, using update steps of $\pm1\%$.
Empirically, we found that using too many covariates as features harms the performance of the MRF model, likely due to the higher dimension of the associated feature space. The best performance, reported in the Results Section, are obtained with only three covariates: building count, average building size, and night lights.

\vspace{1em}
\noindent
\underline{\itshape Convolutional Neural Network (CNN):} It has been proposed to estimate population counts with a CNN with spatial context~\cite{jacobs2018weakly}, trained with the same form of aggregated supervision as \textsc{Pomelo}. We instead advocate for a shared per-pixel architecture. Hence, we further include a baseline that has the same network architecture as \textsc{Pomelo}, with two exceptions: the kernel size is set to 3$\times$3 for all convolutional layers, and the network is trained to directly output population as suggested by~\cite{jacobs2018weakly} (rather than densities, as in our system).
We point out that, while our CNN baseline is inspired by that study, the two are not directly comparable: Jacobs \textit{et al.} target local population mapping in urban areas and base their estimates on high-resolution Planet imagery with 3$\,$m GSD, and on city block-level population and housing counts from the US census. Consequently, they can also afford to train a much larger (U-net~\cite{ronneberger2015u}) model. 

\subsection*{Evaluation metrics}

%\john{}{We could remove this subsection} 
For all three evaluation strategies, shown in the Results section, the estimated maps are evaluated by aggregating the per-pixel counts back to a list of $n_c$ population numbers $\mathbf{\hat{c}}$ at finest available census level and comparing them to the actual census counts $\mathbf{c}$ in terms of $R^2$, mean absolute error (MAE), and mean absolute percentage error (MAPE):

\begin{equation}
\begin{split}
& R^2(\mathbf{\hat{c}},\mathbf{c}) = 1 -\frac{\sum_{j=1}^{n_c}(c_j-\hat{c}_j)^2}{\sum_{j=1}^{n_c}(c_j- \Bar{c})^2}\\
&\text{MAE}(\mathbf{\hat{c}},\mathbf{c}) = \frac{1}{n_c}\sum_{j=1}^{n_c}|c_j-\hat{c}_j|\\
    &\text{MAPE}(\mathbf{\hat{c}},\mathbf{c}) = 100\cdot\frac{1}{n_c}\sum_{j=1}^{n_c}\bigg|\frac{c_j-\hat{c}_j}{c_j}\bigg|
    \end{split}
\end{equation}

%\subsection*{Design choices: loss function, prediction target, data augmentation}

\section*{Data Availability}

All data used in this study is publicly available.
%The population maps as well as the code repository will be made available after publication.
The population maps can be accessed via \href{https://doi.org/10.6084/m9.figshare.21444282.v1 }{https://doi.org/10.6084/m9.figshare.21444282.v1}, while the code is available in the following repository: \href{https://github.com/jvargasmu/population_estimation}{https://github.com/jvargasmu/population\_estimation}.

\section*{Acknowledgments}
This research is funded by the Science and Technology for Humanitarian Action Challenges (HAC) project.

\section*{Author contributions}
D.T., K.S., T.W., and F.O. conceived the study, N.M., J.V., R.D., K.S., and D.T. designed the research, N.M., and J.V. implemented the methods. All the authors discussed the results and wrote the manuscript.

\section*{Competing interests}
The authors declare no competing interests.

\bibliography{bib}

\end{document}